%% file: main_arxiv.tex
\documentclass{article}
\PassOptionsToPackage{numbers,sort&compress}{natbib}
\usepackage[preprint]{neurips_2026}
\usepackage[utf8]{inputenc}
\usepackage[T1]{fontenc}
\usepackage{amsmath,amssymb,amsfonts,bm}
\usepackage{graphicx,booktabs,multirow,makecell,array}
\usepackage{xspace,microtype,url,xcolor}
\usepackage[ruled,vlined,linesnumbered]{algorithm2e}
\SetAlgoNlRelativeSize{-1}
\usepackage{hyperref}
\definecolor{paperblue}{HTML}{0072B2}
\hypersetup{
  colorlinks=true,
  linkcolor=paperblue,
  citecolor=paperblue,
  urlcolor=paperblue,
  pdftitle={UniPose9D: Universal Category-Agnostic Object Pose Estimation},
  pdfauthor={Yang You, Yi Du, Cole Harrison, Leonidas Guibas},
  pdfsubject={Category-agnostic 9D object pose estimation},
  pdfkeywords={object pose estimation, category-agnostic, RGB-D, NOCS, flow matching}
}
\DeclareUnicodeCharacter{2013}{--}
\DeclareUnicodeCharacter{2014}{---}
\DeclareUnicodeCharacter{2019}{'}
\DeclareUnicodeCharacter{00B0}{\ensuremath{^\circ}}
\DeclareUnicodeCharacter{2191}{\ensuremath{\uparrow}}
\DeclareUnicodeCharacter{2193}{\ensuremath{\downarrow}}
\newcommand{\method}{UniPose9D\xspace}
\title{UniPose9D: Universal Category-Agnostic Object Pose Estimation}
\input{arxiv_authors}
\date{}
\begin{document}
\maketitle
\input{sec/0_abstract}
\begin{figure}[h]
  \centering
  \includegraphics[width=.48\linewidth]{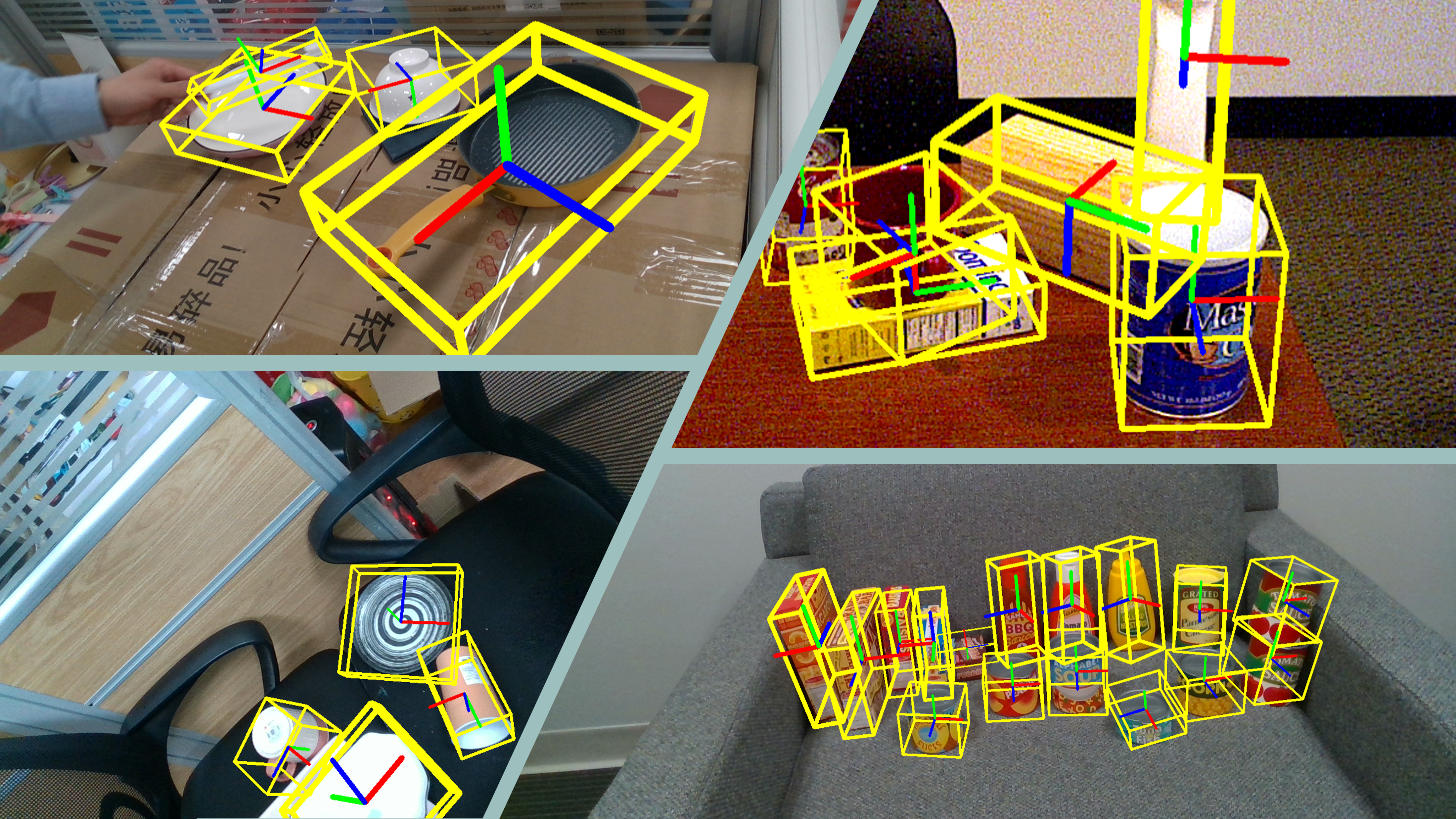}\hfill
  \includegraphics[width=.48\linewidth]{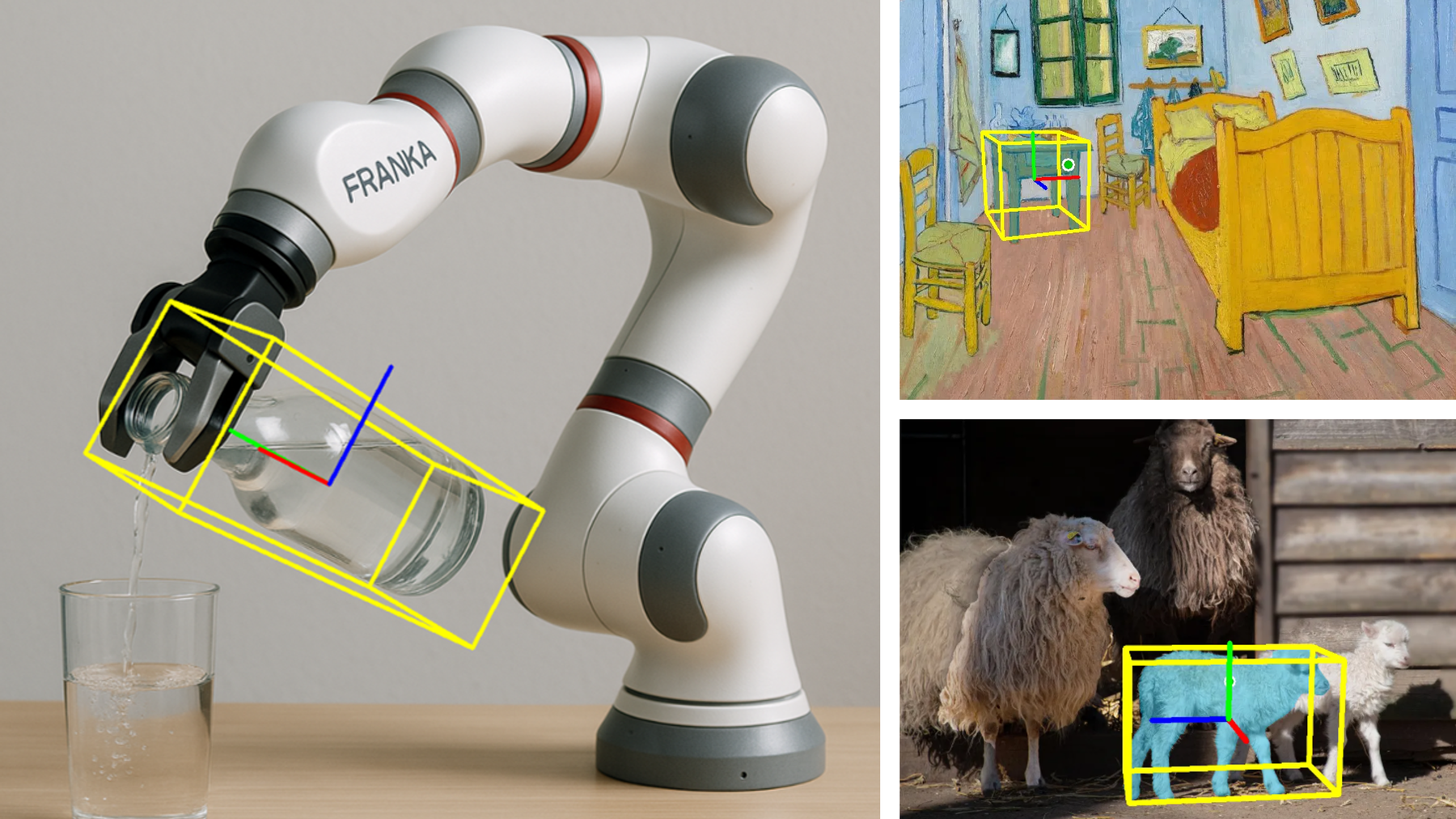}
  \caption{\textbf{One model for category-agnostic 9D pose estimation.} Given an instance mask/ROI and measured depth (left four examples) or predicted depth (right three), \method estimates rotation, translation, and metric box size. No category prompt, target mesh, mean shape, or reference view is supplied at inference. Yellow cuboids show predicted boxes; red, green, and blue axes show orientation.}
  \label{fig:teaser}
\end{figure}
\input{sec/1_intro}
\input{sec/2_related}

\input{sec/3_method}
\input{sec/4_exp}
\input{sec/5_conclusion}
\label{maintextend}
\input{sec/6_acknowledgements}
\bibliographystyle{plainnat}
\bibliography{main}
\clearpage
\appendix
\section*{Supplementary Material}
\setcounter{figure}{0}
\setcounter{table}{0}
\renewcommand{\thefigure}{S\arabic{figure}}
\renewcommand{\thetable}{S\arabic{table}}
\renewcommand{\theHfigure}{supp.\arabic{figure}}
\renewcommand{\theHtable}{supp.\arabic{table}}
\input{sec/review_experiments}
\input{sec/X_suppl}
\end{document}

%% file: arxiv_authors.tex
\author{%
  \makebox[0.44\textwidth][c]{Yang You}\\
  Stanford University\\
  \texttt{yangyou@stanford.edu}\\
  \And
  \makebox[0.44\textwidth][c]{Yi Du}\\
  Stanford University\\
  \texttt{duyi@stanford.edu}\\
  \AND
  \makebox[0.44\textwidth][c]{Cole Harrison}\\
  Amazon\\
  \texttt{colha@amazon.com}\\
  \And
  \makebox[0.44\textwidth][c]{Leonidas Guibas}\\
  Stanford University\\
  \texttt{guibas@cs.stanford.edu}%
}

%% file: sec/0_abstract.tex
\begin{abstract}
Object pose estimation is a fundamental problem in 3D vision. Although recent state-of-the-art approaches achieve strong performance, generalization to novel categories and unseen scenes remains challenging. We propose \textbf{\method}, a unified model for category-agnostic 9D object pose estimation: given an instance mask/ROI and either an RGB-D observation or an RGB image with predicted depth, the model estimates rotation, translation, and metric size without category labels, CAD models, mean-shape priors, or reference views. Specifically, \method samples point pairs from the observed object geometry and uses DINOv2 and PointNet features to predict NOCS coordinates for each pair. To improve accuracy, we introduce a point-pair-based RANSAC \emph{N-hop} Kabsch--Umeyama algorithm with an adaptive threshold. We further employ flow matching to address symmetric ambiguities and construct a large-scale training set by curating and aligning pose annotations from existing public datasets. Experiments across eight datasets show that a single unified model achieves competitive performance on standard benchmarks while generalizing to unseen objects, unseen categories, and in-the-wild scenarios. Our code and model are available at \url{https://github.com/qq456cvb/UniPose9D}.
\end{abstract}

%% file: sec/1_intro.tex
\section{Introduction}
\label{sec:intro}

Object pose estimation is a fundamental problem in computer vision with wide impact on robotic manipulation~\citep{wong2017segicp,wen2022you,zeng2017multi,du2021vision} and augmented reality~\citep{marder2016project,runz2018maskfusion,marchand2015pose}. Traditional instance-level methods~\citep{haugaard2022surfemb,xiang2017posecnn,Wang_2021_GDRN} are tailored to specific objects and typically rely on textured CAD models to synthesize training data, limiting their generalization to unseen instances at test time. Category-level approaches~\citep{wang2019normalized,zheng2023hs,lin2022sar,lin2021dualposenet,chen2021sgpa} remove the need for instance-wise training and meshes, yet they are usually trained and evaluated within predefined categories, and some additionally rely on category-specific models or mean-shape priors.

To mitigate these limitations, recent works aim to estimate the pose of arbitrary objects at inference time by providing a reference 3D textured model or a set of reference images. While this setting enables generalization to unseen objects, it still requires scanning the entire object or carefully selecting reference viewpoints, which is impractical under occlusion and in robotic scenarios where the camera cannot freely move around the target.

This motivates us to ask: do we truly need instance-level meshes/reference views or category labels (e.g., a category name or template)? We call a model \emph{category-agnostic} if, at inference time, it estimates 9D pose for a detected object without receiving its category name, a category template, a mean shape, an instance mesh, or reference views. 
As discussed in Orient Anything~\citep{wang2024orient}, many objects have a naturally defined canonical orientation aligned with human perception, up to rotational symmetries. Consequently, a single observation can provide appearance and geometry cues to predict object pose. We learn these canonical conventions from aligned pose annotations while sharing the same prediction head across categories.

Therefore, we present \textbf{\method}, a universal model that predicts an object's 9D pose, including rotation, translation, and metric bounding box size, from the input described above. Our method goes beyond Orient Anything, which predicts rotation only.

Inspired by CPPF~\citep{you2022cppf}, we avoid per-pixel NOCS~\citep{wang2019normalized} regression. Instead, we sample point pairs from the observed object point cloud, extract trainable geometric features with PointNet~\citep{qi2017pointnet} and visual features with DINOv2~\citep{oquab2023dinov2}, and predict the NOCS coordinates for each pair. Pairwise sampling provides a \emph{quadratic} space of candidate pairs, from which we sample a fixed budget to improve the robustness of the subsequent RANSAC-based Kabsch--Umeyama~\citep{LawrenceBernalWitzgall2019} estimation. For metric-scale recovery, rather than deriving scale directly from Kabsch--Umeyama, we first calibrate it using the ratio between predicted NOCS pair differences and input point-pair differences. We further introduce an \emph{N-hop} Kabsch--Umeyama algorithm that iteratively refines the alignment under a scale-aware inlier threshold. To handle object symmetries, we replace explicit symmetry enumeration with flow matching to model multimodal NOCS distributions.

\textbf{\method} is trained as a single model on a mixture of public pose datasets and attains strong performance on standard benchmarks while generalizing to unseen categories and in-the-wild scenes. We study whether one category-agnostic model can provide a reusable 9D pose estimator across datasets without dataset-specific tuning. Additional evaluation on unseen-category subsets of Omni6D and HANDAL extends our transfer experiments beyond YCB-Video and HOPE.
Our model also accepts RGB input by using predicted metric depth from an off-the-shelf depth estimation model. Our contributions are summarized as follows:
\begin{itemize}
\item We introduce \textbf{\method}, a unified framework that predicts rotation, translation, and metric scale from a single masked RGB-D observation or RGB with predicted depth, without category labels, CAD models, mean-shape priors, or reference images at inference.
\item We propose a point-pair-driven NOCS prediction pipeline, pairwise metric-scale calibration, and an \emph{N-hop} Kabsch--Umeyama scheme with an adaptive inlier threshold, improving robustness and accuracy.
\item Experiments across eight datasets show that \method achieves competitive performance on standard benchmarks and generalizes to unseen categories and in-the-wild images.
\end{itemize}

%% file: sec/2_related.tex
\section{Related Work}

\subsection{Instance-Level Pose Estimation}
Instance-level methods assume access to a CAD model for each target object. Classical pipelines such as PPF~\citep{drost2010model} use depth geometry for matching. Modern systems learn pose from RGB or RGB-D: PoseCNN~\citep{xiang2017posecnn} learns 6D pose, CosyPose~\citep{labbe2020cosypose} integrates coarse-to-fine refinement, SurfEmb~\citep{haugaard2022surfemb} learns dense surface embeddings for 2D--3D matching, and GDR-Net~\citep{Wang_2021_GDRN} uses geometry-aware regression. These methods are strong when instance models are available, but remain tied to target-specific CAD supervision.

\subsection{Category-Level and Category-Agnostic Pose Estimation}
Category-level pose estimation removes the requirement of instance-specific meshes and aims to predict 9D pose for unseen objects within a category. NOCS~\citep{wang2019normalized} introduces a shared canonical space and recovers pose via alignment between predicted coordinates and observed geometry. Many works then extend the paradigm with stronger priors or learning objectives: SGPA~\citep{chen2021sgpa} adapts structure-guided priors, SAR-Net~\citep{lin2022sar} performs shape alignment to handle symmetries and scale, and HS-Pose~\citep{zheng2023hs} designs an operator for extracting hybrid-scope features on point clouds. Alternative formulations target prior-free generalization. VI-Net~\citep{lin2023vi} decouples viewpoint and in-plane rotations to stabilize orientation prediction. IST-Net~\citep{liu2023net} learns an implicit transformation that aligns camera-space features to a canonical frame without category priors. CPPF~\citep{you2022cppf} and CPPF++~\citep{you2022cppf++} use point-pair voting for category-level pose, but their reported models are still trained and evaluated in category-specific settings. Diffusion and generative modeling have also been explored, for example, GenPose~\citep{zhang2023genpose} and GenPose++~\citep{zhang2024omni6dpose}, to model uncertainty and improve robustness.

SecondPose~\citep{chen2024secondpose} combines semantic and geometric features, while CleanPose~\citep{cleanpose2025} improves category-level estimation through causal learning and knowledge distillation. \method shares the broad correspondence-and-alignment philosophy of these methods. It combines point-pair NOCS prediction with pairwise metric-scale calibration and an adaptive N-hop Kabsch--Umeyama solver, using one model trained across heterogeneous public datasets without category labels or category-specific heads at inference.

\subsection{Reference-Based and Open-Vocabulary Pose Estimation}
Reference-based methods avoid category-specific training by using images or videos of the target object at inference. Gen6D~\citep{liu2022gen6d} estimates pose from calibrated reference images, while OnePose~\citep{sun2022onepose} and OnePose++~\citep{he2022onepose++} reconstruct sparse 3D geometry from multi-view inputs using structure-from-motion, then solve pose via 2D--3D matching. FS6D~\citep{he2022fs6d} and FoundationPose~\citep{wen2024foundationpose} extend the idea to RGB-D and achieve strong results when reliable depth and camera poses are available. To reduce data acquisition requirements, recent work explores single-reference or partial-overlap settings. NOPE~\citep{nguyen2024nope} estimates relative orientation from a single anchor image. GigaPose~\citep{nguyen2024gigapose} matches rendered templates of target objects, while Any6D~\citep{lee2025any6d} uses image-to-3D generation to obtain geometric cues from a single reference. These methods introduce a target-specific reference image, video, reconstruction, or 3D model at inference. \method targets the complementary setting in which only the current masked RGB-D observation, or RGB with predicted depth, is available.

\paragraph{Open-vocabulary pose estimation}
Oryon and Horyon~\citep{corsetti2024oryon,corsetti2024horyon} estimate relative 6D pose from two RGB-D observations and object text, while ConceptPose~\citep{kuang2026conceptpose} uses category-derived concepts for cross-view matching. OV9D~\citep{cai2024ov9d} estimates absolute pose and size from a localized RGB-D observation and category description. Pov9D~\citep{wang2025pov9d} and ProM-Pose~\citep{li2026prompose} use language-conditioned shape priors, with ProM-Pose also using temporal observations. Orient Anything V2~\citep{wang2025orientv2} extends orientation estimation to relative rotation and symmetry. \method estimates absolute 9D pose from one localized observation without a category prompt, target reference, or supplied shape prior.

%% file: sec/3_method.tex
\section{Method}
\label{sec:method}

Given a masked RGB-D observation or RGB with predicted metric depth and camera intrinsics, \method back-projects the masked depth to an object point cloud. It predicts point-pair NOCS coordinates and aligns them with observed 3D points to recover rotation $\mathbf{R}\in SO(3)$, translation $\mathbf{t}\in\mathbb{R}^3$, and metric box size $\mathbf{s}\in\mathbb{R}^3$.

\subsection{Preliminary: NOCS}
NOCS canonicalizes each object inside a unit cube by uniformly scaling its CAD model so the longest side of its tight bounding box has length 1, centering it, and consistently aligning objects by center and orientation within each category. These canonical annotations provide training supervision; no category label or target CAD model is needed at inference. Given predicted NOCS coordinates, rotation and translation can be estimated with Kabsch--Umeyama. A NOCS coordinate $\mathbf{x}$ relates to its corresponding 3D point $\mathbf{p}$:
\begin{align}
    \mathbf{p} = \gamma\mathbf{R}\mathbf{x} + \mathbf{t},
\end{align}
where $\gamma$ is the scalar normalization equal to the longest side of the metric 3D bounding box, and the final anisotropic metric box size is denoted by $\mathbf{s}\in\mathbb{R}^3$.

\begin{figure}
    \centering
    \includegraphics[width=\linewidth]{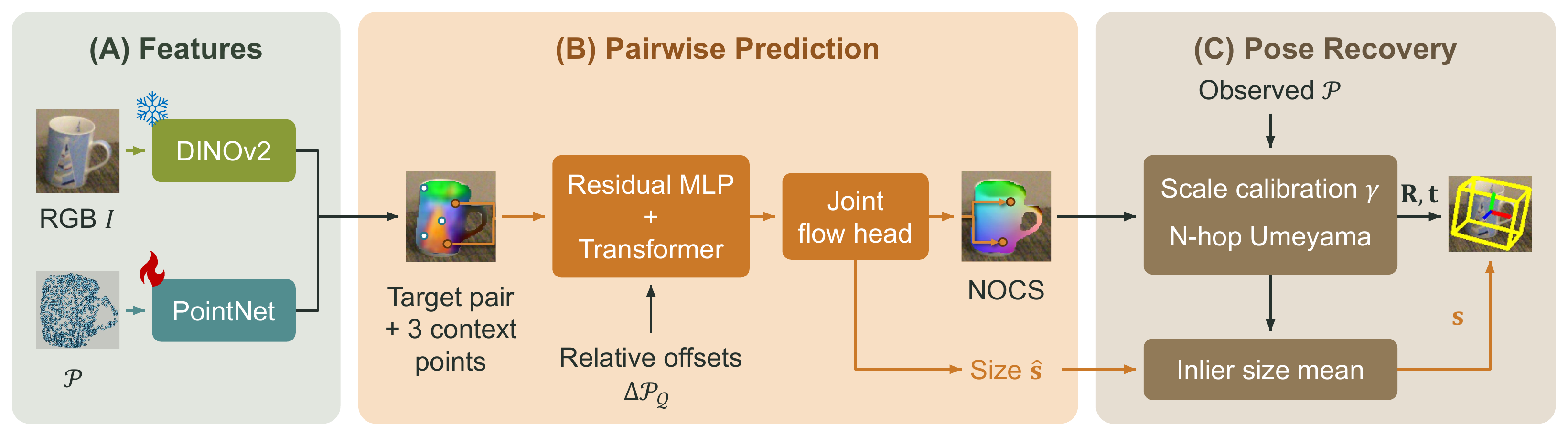}
    \caption{\textbf{Pipeline of \method.} (A) DINOv2 and PointNet extract visual and geometric descriptors. (B) Each target pair is augmented with context points; an MLP and transformer encode their descriptors and relative coordinates. A joint flow head predicts the target pair's NOCS coordinates $\hat{\mathbf{x}}_{i_1},\hat{\mathbf{x}}_{i_2}$ and metric box size $\hat{\mathbf{s}}$. (C) We calibrate scale from pair differences and run N-hop Umeyama with an adaptive threshold to recover $(\mathbf{R},\mathbf{t})$; averaging box sizes over inlier pairs gives $\mathbf{s}$.}

    \label{fig:pipeline}
\end{figure}
\subsection{Overview}

Figure~\ref{fig:pipeline} shows the three parts of \method: visual and geometric feature extraction, point-pair NOCS and box-size prediction, and N-hop Kabsch--Umeyama pose recovery with an adaptive inlier threshold.

\subsection{Feature Extraction}
For visual feature extraction, we use DINOv2-S/14~\citep{oquab2023dinov2} enhanced by 3DCorrEnhance~\citep{you2024multiview}, which provides semantic correspondences across different objects and backgrounds. Given an ROI crop and the image location of each observed point, we bilinearly sample its visual descriptor. For geometric features, a lightweight, multi-radius PointNet aggregates local neighborhoods using residual MLPs and max pooling. We project both modalities to a common width $d$ and concatenate them as $\mathbf{f}_i=[\mathbf{W}_v\mathbf{f}^{\mathrm{img}}_i\oplus\mathbf{W}_g\mathbf{f}^{\mathrm{geo}}_i]\in\mathbb{R}^{2d}$. Full feature-extraction equations and architecture details are given in Appendix~\ref{sec:feature_details}.

\subsection{Point-Pair Sampling and NOCS Prediction}
\paragraph{Point pair construction}
We uniformly sample target pairs $\mathcal{T}{=}(i_1,i_2)$ and augment each with three context points, forming a five-point tuple $\mathcal{Q}_{\mathcal{T}}$. We combine all ten pairwise differences $\Delta\mathcal{P}_{\mathcal{Q}}$ in metric coordinates with projected visual and geometric descriptors:

\begin{align}
    \mathbf{f}_\mathcal{T} = \Delta\mathcal{P}_{\mathcal{Q}} \oplus \mathbf{W}_{p}\!\left(\bigoplus_{j\in\mathcal{Q}_{\mathcal{T}}}\mathbf{f}_j\right).
\end{align}

Five residual MLP blocks followed by a four-layer transformer produce contextual pair features $\mathbf{h}_{\mathcal{T}}$, with attention across the sampled tuples. Architecture dimensions are given in Appendix~\ref{sec:network_details}.

\paragraph{Why point pairs}
Switching from pixel-based NOCS to point-pair-based NOCS offers three advantages: (1) \emph{Quadratic candidate space.} From $N$ sparse samples we obtain $O(N^2)$ possible point pairs, from which we sample $T$ pairs for NOCS prediction and RANSAC. These pairs provide different contexts rather than independent new observations; (2) \emph{Relative invariance.} Pair differences remove common translation biases and make scale calibration easier (Section~\ref{sec:umeyama}); (3) \emph{Occlusion robustness.} Pairwise correspondences allow RANSAC to reject inconsistent hypotheses while retaining geometrically consistent pairs under partial visibility.

\paragraph{Joint flow-matching head}
The head jointly models two NOCS points and metric box size, with target $\mathbf{z}_0=[2\mathbf{x}_{i_1}^{*}\oplus2\mathbf{x}_{i_2}^{*}\oplus\log\mathbf{s}^{*}]\in\mathbb{R}^{9}$. A time-conditioned Gaussian-mixture flow~\citep{gmflow} models multimodal correspondence distributions arising from object symmetries. All nine coordinates follow the linear-noise marginals

\begin{align}
    \mathbf{z}_t = (1-t)\mathbf{z}_0+t\boldsymbol\epsilon, \quad t\in[0,1), \quad \boldsymbol\epsilon\sim\mathcal{N}(0,\mathbf{I}_9).
\end{align}

Training samples a bridged pair $(\mathbf{z}_{t_{\mathrm{low}}},\mathbf{z}_t)$ with $t_{\mathrm{low}}<t$. Conditioned on the contextual feature, noisy state, and learned time embedding $\phi(t)$, the head predicts nine-dimensional flow means, a shared standard deviation, and mixture weights:
\begin{align}
    \{\mathbf{u}_g,\sigma_u,\pi_g\}_{g=1}^G = \operatorname{GM}_\theta(\mathbf{h}_{\mathcal{T}},\mathbf{z}_t,\phi(t)).
\end{align}

These outputs are mapped analytically to reverse-transition means $\mathbf{m}_g$ and variance $v$, as detailed in Appendix~\ref{sec:joint_flow_training}. We minimize the joint negative log-likelihood, averaged over images and tuples:

\begin{align}
    \mathcal{L}_{\mathrm{joint}} = -\log\sum_{g=1}^G\pi_g\mathcal{N}(\mathbf{z}_{t_{\mathrm{low}}}\mid\mathbf{m}_g,v\mathbf{I}_9).
\end{align}

This objective supervises both NOCS and log-size. At test time, an ODE-based sampler produces the joint state; dividing its first six coordinates by two gives the predicted NOCS pair.

\paragraph{Metric box-size output}
Exponentiating the final three sampled coordinates gives the metric box size $\hat{\mathbf{s}}\in\mathbb{R}^3$, learned through the same joint likelihood.
This vector scale is distinct from the scalar NOCS normalization factor $\gamma$ used during correspondence alignment.

\subsection{9D Pose Recovery via N-hop Kabsch--Umeyama}
\label{sec:umeyama}

\paragraph{Scale calibration}
Let $\mathcal{C}{=}\{\mathcal{C}_{\mathcal{T}}\}_{\mathcal{T}}$ collect the predicted NOCS--metric correspondences, with $\mathcal{C}_{\mathcal{T}}{=}\{(\hat{\mathbf{x}}_{i_1},\mathbf{p}_{i_1}),(\hat{\mathbf{x}}_{i_2},\mathbf{p}_{i_2})\}$ for $\mathcal{T}{=}(i_1,i_2)$. Point-pair sampling lets us decouple scalar normalization from Kabsch--Umeyama. For each pair, we calibrate $\gamma$ by
\begin{align}
\label{eq:scale_cal}
    \gamma_\mathcal{T} = \frac{\|\mathbf{p}_{i_2} - \mathbf{p}_{i_1}\|_2}{\|\hat{\mathbf{x}}_{i_2}- \hat{\mathbf{x}}_{i_1}\|_2},
\end{align}

and replace the original coordinates with scaled NOCS $\hat{\mathbf{x}}^{s} = \gamma_\mathcal{T}\hat{\mathbf{x}}$, so that $\mathbf{p}_i = \mathbf{R}\hat{\mathbf{x}}_i^s + \mathbf{t}$. We then apply Kabsch--Umeyama to estimate $(\mathbf{R},\mathbf{t})$ without estimating scale.

Finally, we form a flattened correspondence list by taking both points of every pair and running RANSAC to solve the optimal $SE(3)$ transformation between the scaled NOCS and metric spaces.

\paragraph{Adaptive threshold and N-hop refinement}
We iterate the above RANSAC process for $H{=}3$ hops. Starting from an initial inlier threshold $\tau_0$, at each hop we:
\begin{enumerate}
    \item Run RANSAC--Umeyama at $\tau_h$ to obtain $(\hat{\mathbf{R}},\hat{\mathbf{t}})$ and point-level inliers.
    \item If more than $10\%$ of the individual correspondences are inliers, update the threshold using the largest dimension of the mean box-size vector over selected pairs:
    \begin{align}
        \label{eq:adaptive_threshold}
        \tau_{h+1} = 0.05 \cdot\left\|\frac{1}{|\mathcal{I}|}\sum_{\mathcal{T}_i\in\mathcal{I}}\hat{\mathbf{s}}_{\mathcal{T}_i}\right\|_\infty,
    \end{align}
    where $\mathcal{I}$ contains pairs with at least one inlier endpoint; otherwise, set $\tau_{h+1}=1.5\cdot\tau_h$.
\end{enumerate}

Finally, we average box sizes $\hat{\mathbf{s}}$ over the retained pairs, counting each pair once. Pseudocode is provided in the supplementary material.

\subsection{Implementation Details}
We train one model on the PACE, Omni6DPose, NOCS, and HouseCat6D training splits with balanced sampling and minimize $\mathcal{L}_{\mathrm{joint}}$ using AdamW. At inference, we sample $N=1{,}024$ points and $T=10{,}000$ pairs by default, with 16 ODE steps and 2 substeps per step. Full architecture and optimization settings are in Appendix~\ref{sec:network_details}; Section~\ref{sec:ablation} examines the pair budget.

%% file: sec/4_exp.tex
\section{Experiments}
\label{sec:experiments}
We evaluate a single unified \method model on the held-out test sets of PACE~\citep{you2024pace}, Omni6DPose~\citep{zhang2024omni6dpose}, NOCS REAL275~\citep{wang2019normalized}, and HouseCat6D~\citep{jung2024housecat6d}; the training splits of these four datasets contribute to our training mixture. We then test cross-domain generalization on YCB-Video~\citep{xiang2017posecnn} and HOPE~\citep{tyree2022hope}, evaluate unseen categories in Omni6D~\citep{omni6d2024} and HANDAL~\citep{handal2023}, and report ablations.

\subsection{Experimental Setup}
\label{sec:exp_setup}
We evaluate without test-dataset fine-tuning, using each dataset's standard protocol. \method takes an instance mask/ROI and back-projects measured depth within the mask, or predicted metric depth for RGB input. Appendix~\ref{sec:mask_protocols} summarizes the benchmark mask/ROI conventions.

\paragraph{Baselines and metrics}
We tag methods as \textbf{Cat} (category-level specialist, trained for a fixed taxonomy with category-specific inputs or branches) or \textbf{Gen} (generalizable, requiring no category labels, mean shapes, CAD models, or reference views at inference). Specialist results retain their native training protocols; \method uses one shared model across all four benchmarks. We also retrain baselines on our four-dataset mixture in Appendix~\ref{sec:review_experiments}.

We retain each benchmark's native metrics: PACE reports 3D-box AP at IoU$_{25/50}$ and AP averaged over rotation/translation thresholds; Omni6DPose reports AUC for box overlap and VUS for joint pose thresholds; REAL275 and HouseCat6D report pose mAP. For YCB-Video and HOPE, we report median rotation error and accuracy within $5^\circ$ or 2\,cm. The RGB setting uses metric depth predicted by MoGe-2~\citep{wang2025moge}. Baseline scores are taken from the cited papers or reproduced under the corresponding protocol; metric definitions and additional comparisons are provided in the appendix.

\subsection{Pose Estimation In-Domain}

\paragraph{PACE}
Table~\ref{tab:merged_nocs} summarizes results on the PACE test split. \method attains the best IoU$_{50}$ (\textbf{40.1}) by a wide margin and leads every reported AP metric, with especially large gains over GenPose++ on the joint thresholds AP@0:20$^\circ$,0:5\,cm (\textbf{+20.7}) and AP@0:60$^\circ$,0:15\,cm (\textbf{+22.1}), as well as the pure-angle APs AP@0:20$^\circ$ (\textbf{+18.6}) and AP@0:60$^\circ$ (\textbf{+17.7}). Qualitative comparisons against NOCS, HS-Pose, CPPF++, and GenPose++ are shown in Figure~\ref{fig:pace-vis}.

\begin{table}[!ht]
\centering
\caption{\textbf{PACE benchmark.} 3D-box AP and threshold-averaged pose AP (\%). Cat: category-level specialist; Gen: generalizable. Bold marks the best score.}
\label{tab:merged_nocs}
\resizebox{\linewidth}{!}{
\begin{tabular}{lccccccccc}
\toprule
\multirow{2}{*}{Method} & \multirow{2}{*}{Type} &
\multirow{2}{*}{IoU$_{25}\uparrow$} & \multirow{2}{*}{IoU$_{50}\uparrow$} &
\multicolumn{6}{c}{AP $\uparrow$} \\
\cmidrule(lr){5-10}
 & & & & 0:20$^\circ$ & 0:60$^\circ$ & 0:5\,cm & 0:15\,cm & 0:20$^\circ$,\,0:5\,cm & 0:60$^\circ$,\,0:15\,cm \\
\midrule
NOCS~\citep{wang2019normalized}            & Cat  & 0.2  & 0.0  & 1.2  & 4.3  & 17.0 & 33.5 & 1.1  & 4.2  \\
HS-Pose~\citep{zheng2023hs}                & Cat  & 36.6 & 2.7  & 5.3  & 8.6  & 48.7 & 81.0 & 3.9  & 8.1  \\
SGPA~\citep{chen2021sgpa}                  & Cat  & 2.6  & 1.2  & 5.6  & 11.4 & 16.1 & 50.7 & 3.3  & 10.1 \\
DualPoseNet~\citep{lin2021dualposenet}     & Cat  & 24.2 & 0.1  & 5.5  & 8.1  & 18.6 & 63.4 & 1.8  & 6.5  \\
SAR-Net~\citep{lin2022sar}                 & Cat  & 22.8 & 0.2  & 5.3  & 8.8  & 48.6 & 79.8 & 3.7  & 8.2  \\
CPPF++~\citep{you2022cppf++}               & Cat  & 44.5 & 4.4  & 15.2 & 27.3 & 35.3 & 74.0 & 9.9  & 24.9 \\
GenPose++~\citep{zhang2024omni6dpose}      & Gen & 52.5 & 18.6 & 21.8 & 39.7 & 52.3 & 79.0 & 15.7 & 33.9 \\
\midrule
Ours & Gen & \textbf{61.7} & \textbf{40.1} & \textbf{40.4} & \textbf{57.4} & \textbf{65.2} & \textbf{87.7} & \textbf{36.4} & \textbf{56.0} \\
\bottomrule
\end{tabular}
}
\end{table}

\paragraph{Omni6DPose}
Table~\ref{tab:pose_comparison} shows that \method ranks first across AUC and VUS, surpassing GenPose++ on all three AUC@IoU thresholds while leading or tying on every VUS metric. The improvement is larger for 3D-box overlap (AUC@IoU$_{50}$: 19.1 to 23.0) than for the joint pose metrics (VUS@$5^\circ$5\,cm: 15.1 to 15.8).

\begin{table}[!ht]
\centering
\caption{\textbf{Omni6DPose benchmark.} AUC for box overlap and VUS for joint pose thresholds (\%). The same \method model is used across all four in-domain datasets.}
\label{tab:pose_comparison}
\setlength{\tabcolsep}{5pt}
\resizebox{\linewidth}{!}{
\begin{tabular}{lcccccccc}
\toprule
\multirow{2}{*}{Method} & \multirow{2}{*}{Type} &
\multicolumn{3}{c}{AUC $\uparrow$} & \multicolumn{4}{c}{VUS $\uparrow$} \\
\cmidrule(lr){3-5} \cmidrule(lr){6-9}
 & & IoU$_{25}$ & IoU$_{50}$ & IoU$_{75}$ &
 $5^\circ$2\,cm & $5^\circ$5\,cm & $10^\circ$2\,cm & $10^\circ$5\,cm \\
\midrule
NOCS~\citep{wang2019normalized}       & Cat  & 0.0  & 0.0  & 0.0 & 0.0  & 0.0  & 0.0           & 0.0  \\
SGPA~\citep{chen2021sgpa}             & Cat  & 10.5 & 2.0  & 0.0 & 4.3  & 6.7  & 9.3           & 15.0 \\
IST-Net~\citep{liu2023net}            & Cat  & 28.7 & 10.6 & 0.5 & 2.0  & 3.4  & 5.3           & 8.8  \\
HS-Pose~\citep{zheng2023hs}           & Cat  & 31.6 & 13.6 & 1.1 & 3.5  & 5.3  & 8.4           & 12.7 \\
GenPose++~\citep{zhang2024omni6dpose} & Gen & 39.0 & 19.1 & 2.0 & 10.0 & 15.1 & \textbf{19.5} & 29.4 \\
\midrule
Ours & Gen & \textbf{43.1} & \textbf{23.0} & \textbf{2.9} & \textbf{10.5} & \textbf{15.8} & \textbf{19.5} & \textbf{29.7} \\
\bottomrule
\end{tabular}
}
\end{table}

\paragraph{NOCS REAL275}
Table~\ref{tab:real275} compares \method with category-level pose estimation methods. \method attains a competitive IoU$_{75}$ of 48.1, within 1.6 points of SecondPose, and comparable mAP at looser angular thresholds. The recent specialist CleanPose~\citep{cleanpose2025} achieves a higher published $5^\circ$5\,cm mAP (67.6 versus 60.2). Our model uses the same category-agnostic weights across all four benchmarks without mean-shape priors at inference.

\begin{table}[!ht]
\centering
\caption{\textbf{NOCS REAL275 benchmark.} IoU$_{75}$ and pose mAP (\%). Dashes denote unreported values; bold marks the best score.}
\label{tab:real275}
\setlength{\tabcolsep}{8pt}
\resizebox{\linewidth}{!}{
\begin{tabular}{lcccccc}
\toprule
Method & Type & IoU$_{75}$ & $5^\circ$2\,cm & $5^\circ$5\,cm & $10^\circ$2\,cm & $10^\circ$5\,cm \\
\midrule
FS-Net~\citep{chen2021fs}              & Cat  & --   & --   & 28.2 & --   & 60.8 \\
DualPoseNet~\citep{lin2021dualposenet} & Cat  & 30.8 & 29.3 & 35.9 & 50.0 & 66.8 \\
GPV-Pose~\citep{di2022gpv}             & Cat  & --   & 32.0 & 42.9 & --   & 71.6 \\
SS-ConvNet~\citep{lin2021sparse}       & Cat  & --   & 36.6 & 43.4 & 52.6 & 63.5 \\
HS-Pose~\citep{zheng2023hs}            & Cat  & --   & 41.3 & 50.4 & 72.3 & 82.7 \\
IST-Net~\citep{liu2023net}             & Cat  & 47.5 & 53.4 & 63.2 & 72.1 & 80.5 \\
VI-Net~\citep{lin2023vi}               & Cat  & 48.3 & 50.0 & 57.6 & 70.8 & 82.1 \\
SecondPose~\citep{chen2024secondpose}  & Cat  & 49.7 & 56.2 & 63.6 & 74.7 & 86.0 \\
CleanPose~\citep{cleanpose2025} & Cat & \textbf{62.7} & \textbf{61.7} & \textbf{67.6} & \textbf{78.3} & \textbf{86.3} \\
\midrule
Ours & Gen & 48.1 & 45.4 & 60.2 & 63.6 & 82.9 \\
\bottomrule
\end{tabular}
}
\end{table}

\paragraph{HouseCat6D}
Table~\ref{tab:housecat6d} shows strong performance in cluttered household scenes: \method improves $5^\circ$5\,cm mAP from SecondPose's 13.4 to 19.0, while CleanPose reports a higher score of 24.1. These native-protocol comparisons are complemented by retraining baselines on our four-dataset mixture in Table~\ref{tab:all_dataset_baselines}.

\begin{table}[!ht]
\centering
\caption{\textbf{HouseCat6D benchmark.} 3D-box AP and pose mAP (\%) under the native protocol. Bold marks the best score.}
\label{tab:housecat6d}
\setlength{\tabcolsep}{6pt}
\resizebox{\linewidth}{!}{
\begin{tabular}{lccccccc}
\toprule
Method & Type & IoU$_{25}$ & IoU$_{50}$ & $5^\circ$2\,cm & $5^\circ$5\,cm & $10^\circ$2\,cm & $10^\circ$5\,cm \\
\midrule
FS-Net~\citep{chen2021fs}             & Cat  & 74.9 & 48.0 & 3.3  & 4.2  & 17.1 & 21.6 \\
GPV-Pose~\citep{di2022gpv}            & Cat  & 74.9 & 50.7 & 3.5  & 4.6  & 17.8 & 22.7 \\
VI-Net~\citep{lin2023vi}              & Cat  & 80.7 & 56.4 & 8.4  & 10.3 & 20.5 & 29.1 \\
AG-Pose~\citep{lin2024instance}       & Cat  & 81.8 & 62.5 & 11.5 & 12.0 & 32.7 & 35.8 \\
SecondPose~\citep{chen2024secondpose} & Cat  & 83.7 & 66.1 & 11.0 & 13.4 & 25.3 & 35.7 \\
CleanPose~\citep{cleanpose2025} & Cat & \textbf{89.2} & \textbf{79.8} & \textbf{22.4} & \textbf{24.1} & \textbf{51.6} & \textbf{56.5} \\
\midrule
Ours & Gen & 86.4 & 70.6 & 16.9 & 19.0 & 37.5 & 43.2 \\
\bottomrule
\end{tabular}
}
\end{table}

\subsection{Generalization to Unseen Instances}
To test generalization to unseen instances, we evaluate on the full YCB-Video and HOPE test sets. Both contain seen and unseen categories, while all object instances and scenes are absent from our pose-training mixture. We compare with Orient Anything (rotation-only) and GenPose++, calibrating rotations to a common canonical convention. In Table~\ref{tab:in-the-wild}, \method improves RGB-D accuracy over GenPose++ by \textbf{28.0/31.3} percentage points at $5^\circ$ and \textbf{45.4/35.8} points at 2\,cm on YCB-Video/HOPE, respectively, with median rotation errors around $4^\circ$ on both. RGB rotation accuracy also improves. Qualitative comparisons appear in the supplementary material.

With depth predicted by MoGe-2~\citep{wang2025moge}, metric translation remains difficult: Acc@2\,cm falls from 91.9\% to 3.4\% on YCB-Video and from 91.5\% to 1.5\% on HOPE when moving from RGB-D to RGB. On HOPE with RGB input, GenPose++ has higher translation accuracy (2.3\%). Rotation is less affected: our median error changes from $4.0^\circ$ to $8.9^\circ$ on YCB-Video and from $4.2^\circ$ to $7.6^\circ$ on HOPE. Predicted-depth scale errors directly affect metric translation.

\begin{table}[!ht]
\centering
\caption{\textbf{YCB-Video and HOPE cross-domain evaluation.} Median rotation error (Med$^\circ$), Acc@5$^\circ$, and Acc@2\,cm (\%). OriAny predicts rotation only.}
\label{tab:in-the-wild}
\setlength{\tabcolsep}{5pt}
\resizebox{\linewidth}{!}{
\begin{tabular}{lllcccccc}
\toprule
\multirow{2}{*}{Setting} & \multirow{2}{*}{Method} & \multirow{2}{*}{Type} &
\multicolumn{3}{c}{YCB-Video} & \multicolumn{3}{c}{HOPE} \\
\cmidrule(lr){4-6} \cmidrule(lr){7-9}
& & & Med$^{\circ}\!\downarrow$ & 5$^{\circ}\!\uparrow$ & 2\,cm$\uparrow$
& Med$^{\circ}\!\downarrow$ & 5$^{\circ}\!\uparrow$ & 2\,cm$\uparrow$ \\
\midrule
\multirow{3}{*}{RGB}
& OriAny~\citep{wang2024orient}         & Gen & 91.4 & 1.4  & --  & 77.4 & 1.1  & --  \\
& GenPose++~\citep{zhang2024omni6dpose} & Gen & 15.5 & 14.4 & 1.5 & 63.6 & 19.8 & \textbf{2.3} \\
& \textbf{Ours}                        & Gen & \textbf{8.9} & \textbf{27.9} & \textbf{3.4} & \textbf{7.6} & \textbf{33.4} & 1.5 \\
\midrule
\multirow{2}{*}{RGB-D}
& GenPose++~\citep{zhang2024omni6dpose} & Gen & 10.3 & 33.9 & 46.5 & 53.4 & 25.7 & 55.7 \\
& \textbf{Ours}                        & Gen & \textbf{4.0} & \textbf{61.9} & \textbf{91.9} & \textbf{4.2} & \textbf{57.0} & \textbf{91.5} \\
\bottomrule
\end{tabular}
}
\end{table}

\subsection{Generalization to Unseen Categories}
\label{sec:unseen_categories}
\paragraph{YCB-Video and HOPE}
We next evaluate categories absent from our four-dataset pose-training mixture. YCB-Video contains five such categories (3,543 examples), and HOPE contains two (225 examples), with joint $5^\circ$5\,cm accuracies of 40.8\% and 64.9\%, respectively. Category lists and complete subset results are in Appendix~\ref{sec:unseen_details}.

\paragraph{Omni6D}
Omni6D~\citep{omni6d2024} is distinct from the Omni6DPose training dataset. The baselines in Table~\ref{tab:omni6d_unseen} train on Omni6D\_train and test on its held-out categories. \method transfers zero-shot to the same unseen split without Omni6D training, achieving 50.71\% accuracy at IoU$_{50}$ and 26.42\% at $5^\circ$5\,cm on 9,530 examples from 13 categories absent from our pose-training mixture.
\begin{table}[!ht]
\centering
\caption{\textbf{Omni6D unseen-split evaluation.} Scores (\%); baselines train on Omni6D\_train, while ours is zero-shot. Baseline results are from \citet{omni6d2024}; metric definitions and evaluation details are in Appendix~\ref{sec:unseen_details}.}
\label{tab:omni6d_unseen}
\resizebox{\linewidth}{!}{
\begin{tabular}{lccccccccc}
\toprule
\multirow{2}{*}{Method} & \multirow{2}{*}{IoU$_{50}$} & \multicolumn{4}{c}{Joint pose} & \multicolumn{2}{c}{Rotation} & \multicolumn{2}{c}{Translation} \\
\cmidrule(lr){3-6}\cmidrule(lr){7-8}\cmidrule(lr){9-10}
& & $5^\circ$2\,cm & $5^\circ$5\,cm & $10^\circ$2\,cm & $10^\circ$5\,cm & $5^\circ$ & $10^\circ$ & 2\,cm & 5\,cm \\
\midrule
SGPA & 7.05 & 0.07 & 0.28 & 0.19 & 0.82 & 0.53 & 1.69 & 3.87 & 28.28 \\
DualPoseNet & 36.85 & 3.24 & 3.37 & 8.04 & 8.51 & 3.39 & 8.64 & 78.00 & 98.60 \\
HS-Pose & 36.75 & 1.54 & 1.66 & 4.67 & 5.16 & 1.75 & 5.38 & 79.95 & 98.27 \\
\midrule
\textbf{Ours (zero-shot)} & 50.71 & 15.61 & 26.42 & 20.26 & 32.26 & 29.77 & 35.89 & 48.96 & 86.20 \\
\bottomrule
\end{tabular}
}
\end{table}

\paragraph{HANDAL}
We further evaluate five unseen categories in HANDAL~\citep{handal2023}: ladle, power drill, spatula, strainer, and whisk. Table~\ref{tab:handal_diagnostic} compares the same 25 examples for all methods. \method obtains a median rotation error of $3.3^\circ$, compared with $33.9^\circ$ for GenPose++ and $68.1^\circ$ for Orient Anything. Because Orient Anything predicts only rotation, it receives ground-truth translation in this comparison. The sample is small and serves as a diagnostic rather than a full HANDAL benchmark.
\begin{table}[!ht]
\centering
\caption{\textbf{HANDAL unseen-category evaluation.} Joint-threshold scores (\%) and median rotation error on 25 examples from five categories absent from pose training. Orient Anything receives ground-truth translation.}
\label{tab:handal_diagnostic}
\resizebox{\linewidth}{!}{
\begin{tabular}{lrrrrr}
\toprule
Method & $5^\circ$2\,cm & $5^\circ$5\,cm & $10^\circ$2\,cm & $10^\circ$5\,cm & Median rot. ($^\circ$) \\
\midrule
Orient Anything~\citep{wang2024orient} & 0.0 & 0.0 & 8.0 & 8.0 & 68.1 \\
GenPose++~\citep{zhang2024omni6dpose} & 0.0 & 4.0 & 16.0 & 28.0 & 33.9 \\
Ours & \textbf{56.0} & \textbf{64.0} & \textbf{60.0} & \textbf{76.0} & \textbf{3.3} \\
\bottomrule
\end{tabular}
}
\end{table}

\subsection{Ablation Studies}
\label{sec:ablation}

\begin{table}[!h]
\centering
\caption{\textbf{Ablation study on the NOCS REAL275 dataset.} Full model, component ablations, and plain RANSAC.}
\label{tab:real275_ablation}
\small
\begin{tabular}{lccccc}
\toprule
Variant & IoU$_{75}$ & $5^\circ$2\,cm & $5^\circ$5\,cm & $10^\circ$2\,cm & $10^\circ$5\,cm \\
\midrule
Full model (Ours) & \textbf{48.1} & \textbf{45.4} & \textbf{60.2} & \textbf{63.6} & \textbf{82.9} \\
\midrule
w/o PointNet features & 22.9 & 24.5 & 34.2 & 40.0 & 70.3 \\
w/o point-pair sampling & 44.7 & 36.9 & 51.6 & 55.1 & 77.5 \\
w/o flow matching    & 41.5 & 25.9 & 33.2 & 50.1 & 65.1 \\
w/o scale decoupling & 44.5 & 44.0 & 57.5 & 61.2 & 80.5 \\
Plain RANSAC        & 43.4 & 43.7 & 58.2 & 59.7 & 78.9 \\
\bottomrule
\end{tabular}
\end{table}

\paragraph{Component analysis}
Table~\ref{tab:real275_ablation} compares component ablations and plain RANSAC. Excluding geometric PointNet features causes a large drop in IoU$_{75}$ and pose mAP. Removing point-pair sampling, flow matching, or scale decoupling reduces accuracy. Our full recovery also outperforms plain RANSAC. These results support the learned correspondences and geometric recovery; point pairs provide local hypotheses for robust fitting.

\paragraph{Runtime and solver settings}
On one RTX A6000, the reported pose-stage latency is 2.784\,s with 10,000 pairs; 3,000 pairs gives similar REAL275 accuracy at 0.979\,s. Tables~\ref{tab:pair_runtime}--\ref{tab:solver_sensitivity} report pair-count, profiling, and solver-sensitivity results. Three refinement hops remain the default; one to three hops gives $5^\circ$5\,cm mAP of 59.9--60.2.

\paragraph{Mask and depth sensitivity}
We test controlled input corruptions on the full REAL275 test set. Relative to clean-input $5^\circ$5\,cm mAP of 60.2, mask erosion by 10\% of the equivalent object radius gives 57.8, and 10\% depth outliers give 56.6. Dilation and ragged boundaries at the tested severity give 62.9 and 62.1. The perturbation definitions and complete results are in Table~\ref{tab:input_robustness}.

%% file: sec/5_conclusion.tex
\section{Conclusion}
\label{sec:conclusion}

We introduced \method, a universal model for category-agnostic 9D object pose estimation from a masked RGB-D observation or RGB with predicted depth. \method combines point-pair NOCS prediction, pairwise metric-scale calibration, and adaptive N-hop Kabsch--Umeyama recovery. A single model trained across public datasets achieves competitive benchmark performance and transfers to unseen objects and categories without category labels, CAD models, mean-shape priors, or reference views at inference. Ablations confirm the contributions of point-pair reasoning, flow matching, geometric features, and the pose-recovery components.

\paragraph{Limitations}
Our method currently focuses on rigid indoor objects of moderate size and assumes an instance mask/ROI. Its accuracy depends on mask and depth quality; monocular depth errors particularly affect metric translation. Flow sampling dominates inference latency, limiting applications that require frequent pose updates. We plan to incorporate more outdoor objects and datasets and extend \method to articulated and deformable categories.

%% file: sec/6_acknowledgements.tex
\section*{Acknowledgements}

Leonidas Guibas and Yang You acknowledge support from the Toyota Research Institute University 3.0 Program, ARL grant W911NF-21-2-0104, a Vannevar Bush Faculty Fellowship, and a gift from the Flexiv corporation. Yang You is also supported in part by the Outstanding Doctoral Graduates Development Scholarship of Shanghai Jiao Tong University.

%% file: sec/review_experiments.tex
\section{Additional Evaluation Results}
\label{sec:review_experiments}
\paragraph{Native metrics.}
PACE AP@0:20$^\circ$ averages thresholds from $0^\circ$ to $20^\circ$ in $1^\circ$ steps; AP@0:5\,cm uses 0.25\,cm steps. Joint AP evaluates both constraints. Omni6DPose uses its native AUC and VUS protocol, while REAL275 and HouseCat6D use pose mAP. These metrics are retained separately throughout the comparisons.

\subsection{Unified Training of Baselines}
Table~\ref{tab:all_dataset_baselines} compares methods trained on the same four-dataset mixture. We retain the native metric for each benchmark: threshold-integrated AP on PACE, VUS on Omni6DPose, and pose mAP on REAL275 and HouseCat6D. \method leads the jointly trained variants in all four columns. The NOCS-only SecondPose and CleanPose checkpoints retain higher REAL275 accuracy; their scores and those of the jointly trained models correspond to different training settings. CleanPose uses a 190-category head for the combined taxonomy. Its NOCS-only diagnostic score is reproduced, distinct from the published result in Table~\ref{tab:real275}.

\begin{table}[htbp]
\centering
\caption{\textbf{Baselines under the unified four-dataset training protocol.} ``All-DS'' denotes training on the same PACE, Omni6DPose, NOCS REAL275, and HouseCat6D mixture as \method. Type: \textbf{Cat} = category-level specialist, \textbf{Gen} = generalizable. SecondPose and CleanPose remain category-conditioned under All-DS training. CleanPose (All-DS) uses a 190-category head in this diagnostic. Dashes indicate unreported native-checkpoint results, not measured zero scores.}
\label{tab:all_dataset_baselines}
\resizebox{\linewidth}{!}{
\begin{tabular}{llcccc}
\toprule
Method & Type & PACE (0:20$^\circ$) & Omni6DPose (5$^\circ$2\,cm) & REAL275 (5$^\circ$2\,cm) & HouseCat6D (5$^\circ$2\,cm) \\
\midrule
GenPose++ (Omni)    & Gen & 21.8 & 10.0 & 10.3 & 1.8 \\
GenPose++ (All-DS)  & Gen & 31.2 & 7.1  & 28.5 & 9.8 \\
\midrule
SecondPose (NOCS)   & Cat  & --   & --   & 56.2 & -- \\
SecondPose (All-DS) & Cat  & 24.5 & 5.1  & 36.2 & 7.9 \\
CleanPose (NOCS, reproduced) & Cat & -- & -- & \textbf{61.6} & -- \\
CleanPose (All-DS) & Cat & 13.4 & 2.3 & 4.0 & 1.7 \\
\midrule
\method & Gen & \textbf{40.4} & \textbf{10.5} & 45.4 & \textbf{16.9} \\
\bottomrule
\end{tabular}
}
\end{table}

\subsection{Unseen-Category Evaluation Details}
\label{sec:unseen_details}
The category subsets in Table~\ref{tab:unseen_categories} exclude categories represented in the four-dataset pose-training mixture. This does not assert that the same categories were absent from pretrained DINOv2 or MoGe-2 data. We use the category names recorded in the evaluation:
\begin{itemize}
    \item \textbf{YCB-Video (5):} clamp, foam brick, pitcher, power drill, wood block.
    \item \textbf{HOPE (2):} squeeze bottle, milk carton.
    \item \textbf{Omni6D (13):} almond, apricot, artichoke, bagel, chestnut, earplug, fig, ginger, pagoda dish, passiflora edulis, passion fruit, pistachio, taro.
    \item \textbf{HANDAL (5):} ladle, power drill, spatula, strainer, whisk.
\end{itemize}
Each joint threshold requires both rotation and translation errors to fall below the specified limits. The subset scores are reported separately for each dataset. Omni6D is a separate dataset from Omni6DPose. The baselines train on Omni6D\_train and test on the held-out-category split. Our zero-shot evaluation uses 9,530 examples from the listed 13 categories of the same unseen split, without Omni6D pose training.

Published baseline results in Table~\ref{tab:omni6d_unseen} report mAP. Our Omni6D evaluation uses GT visible masks and the released object symmetry annotations, and reports pooled per-instance accuracy. It uses 1,024 input points, 10,000 point pairs, and two RANSAC fitting passes with up to 1,000 hypotheses per pass. The same-sample HANDAL comparison is given in Table~\ref{tab:handal_diagnostic}.

\begin{table}[htbp]
\centering
\caption{\textbf{Category-held-out subsets.} Joint-threshold scores (\%) for RGB-D input. Counts denote evaluated examples. The full Omni6D result and same-subset HANDAL comparison appear in Tables~\ref{tab:omni6d_unseen} and~\ref{tab:handal_diagnostic}.}
\label{tab:unseen_categories}
\small
\setlength{\tabcolsep}{6pt}
\begin{tabular}{lrrrrrr}
\toprule
Dataset & Categories & Count & $5^\circ$2\,cm & $5^\circ$5\,cm & $10^\circ$2\,cm & $10^\circ$5\,cm \\
\midrule
YCB-Video & 5 & 3,543 & 40.6 & 40.8 & 63.8 & 64.2 \\
HOPE & 2 & 225 & 64.9 & 64.9 & 85.8 & 86.7 \\
Omni6D & 13 & 9,530 & 15.6 & 26.4 & 20.3 & 32.3 \\
HANDAL & 5 & 25 & 56.0 & 64.0 & 60.0 & 76.0 \\
\bottomrule
\end{tabular}
\end{table}

\subsection{Runtime and Pair Sampling}
We draw $T$ pairs from $N=1{,}024$ sampled object points. The $O(N^2)$ candidate pool is not exhaustively evaluated. Table~\ref{tab:pair_runtime} varies only the pair count in the recorded REAL275 evaluation, keeping $T=10{,}000$ as the default for the paper's other reported results. Peak GPU memory rises from 1.11 to 1.48\,GiB over the tested range. \begin{table}[htbp]
\centering
\caption{\textbf{Sampled-pair cost on REAL275.} Reported pose-stage latency on one RTX A6000, $N=1{,}024$ object points. All main results retain $T=10{,}000$ pairs; $T=3{,}000$ is an alternative operating point. Module-level latency is in Table~\ref{tab:runtime_breakdown}.}
\label{tab:pair_runtime}
\small
\setlength{\tabcolsep}{8pt}
\begin{tabular}{lrrrr}
\toprule
Pairs $T$ & mAP@$5^\circ$5\,cm & mAP@$10^\circ$5\,cm & Latency (s) & GPU memory (GiB) \\
\midrule
1,000 & 58.4 & 82.2 & 0.837 & 1.11 \\
3,000 & 60.3 & 83.0 & 0.979 & 1.17 \\
10,000 (default) & 60.2 & 82.9 & 2.784 & 1.48 \\
\bottomrule
\end{tabular}
\end{table}

Table~\ref{tab:runtime_breakdown} gives the recorded latency components on one RTX A6000. 
\begin{table}[!htbp]
\centering
\caption{\textbf{Per-object module latency.} Default $T=10{,}000$ on one RTX A6000. These measurements characterize the reported pose pipeline and do not establish end-to-end RGB latency including mask and depth prediction.}
\label{tab:runtime_breakdown}
\small
\begin{tabular}{lr}
\toprule
Component & Time (s) \\
\midrule
Preprocessing and aggregation & 0.105 \\
Point features & 0.117 \\
Flow sampling & 2.360 \\
RANSAC & 0.201 \\
\midrule
Reported total & 2.784 \\
\bottomrule
\end{tabular}
\end{table}

\subsection{Solver Sensitivity}
Table~\ref{tab:solver_sensitivity} evaluates the threshold and refinement-hop settings on the full REAL275 test set. The adaptive threshold is $\tau=0.05\bar{s}_\infty$, where $\bar{s}_\infty=\|\,|\mathcal{I}|^{-1}\sum_{\mathcal{T}\in\mathcal{I}}\hat{\mathbf{s}}_{\mathcal{T}}\|_\infty$ is the largest dimension of the mean box-size vector over selected pairs, as in Equation~\eqref{eq:adaptive_threshold}. The threshold is relaxed by a factor of 1.5 when support is insufficient. The refinement loop is summarized in Algorithm~\ref{alg:nhop}. Fixed 1\,cm and 2\,cm thresholds give similar scores on this dataset; a 5\,cm threshold reduces accuracy. The difference between one and three hops is 0.3 points at $5^\circ$5\,cm. These measurements support limited sensitivity over the tested hop range, but do not by themselves validate threshold transfer across all object sizes.

\begin{table}[!htbp]
\centering
\caption{\textbf{Solver sensitivity on REAL275.} Pose mAP (\%); the default adaptive threshold and three hops are retained.}
\label{tab:solver_sensitivity}
\small
\setlength{\tabcolsep}{12pt}
\begin{tabular}{lrr}
\toprule
Setting & $5^\circ$5\,cm & $10^\circ$5\,cm \\
\midrule
Adaptive threshold, $0.05\bar{s}_\infty$ (default) & 60.2 & 82.9 \\
Fixed threshold, 1\,cm & 59.8 & 82.8 \\
Fixed threshold, 2\,cm & 59.6 & 81.6 \\
Fixed threshold, 5\,cm & 50.3 & 76.4 \\
\midrule
1 refinement hop & 59.9 & 82.7 \\
2 refinement hops & 60.1 & 82.8 \\
3 refinement hops (default) & 60.2 & 82.9 \\
\bottomrule
\end{tabular}
\end{table}

\subsection{Mask and Depth Perturbations}
We evaluate controlled input corruptions on the full REAL275 test set (2,754 images), with $T=10{,}000$ pairs and 1,000 RANSAC hypotheses. For a mask of area $A$, severities are relative to its equivalent radius $r_{\mathrm{eq}}=\sqrt{A/\pi}$. Erosion and dilation use 10\% of this radius. The ragged-boundary test flips each pixel in a band of this width around the silhouette with probability 0.5; this operation does not guarantee identical mask area. The depth test corrupts 10\% of depth values. Table~\ref{tab:input_robustness} reports $5^\circ$5\,cm mAP. Erosion and depth outliers reduce the score by 2.4 and 3.6 points; the other two perturbations yield higher scores in this recorded run. These tests probe specific corruptions and do not cover missing instances, severe truncation, or arbitrary segmentation failures.

\begin{table}[!htbp]
\centering
\caption{\textbf{Controlled input perturbations on REAL275.} Changes are in percentage points relative to clean-input $5^\circ$5\,cm mAP.}
\label{tab:input_robustness}
\small
\setlength{\tabcolsep}{12pt}
\begin{tabular}{lrr}
\toprule
Input condition & mAP@$5^\circ$5\,cm & Change \\
\midrule
Clean & 60.2 & -- \\
Mask erosion, $0.1r_{\mathrm{eq}}$ & 57.8 & $-2.4$ \\
Mask dilation, $0.1r_{\mathrm{eq}}$ & 62.9 & $+2.7$ \\
Ragged boundary, $0.1r_{\mathrm{eq}}$ band & 62.1 & $+1.9$ \\
Depth outliers, 10\% & 56.6 & $-3.6$ \\
\bottomrule
\end{tabular}
\end{table}

\subsection{Component Ablation Definitions}
\label{sec:ablation_definitions}
In Table~\ref{tab:real275_ablation}, ``w/o flow matching'' uses direct regression with an $\ell_2$ loss. ``w/o point-pair sampling'' independently predicts each point's NOCS coordinates from its concatenated PointNet and DINOv2 features. Plain RANSAC directly fits a similarity transform without pairwise scale calibration; $H=1$ in Table~\ref{tab:solver_sensitivity} retains the proposed scale calibration and rigid fitting, changing only the hop count.

%% file: sec/X_suppl.tex
\section{Network Implementation Details}
\label{sec:network_details}
\subsection{Feature Extraction Formulation}
\label{sec:feature_details}
For visual feature extraction, we use DINOv2-S/14 enhanced by 3DCorrEnhance~\citep{you2024multiview}. Given an ROI crop $\mathbf{I}\in\mathbb{R}^{H\times W\times 3}$ and the image location $\mathbf{u}_i$ of a point, we refine the patch-token map with a $3\times3$ convolution, bilinearly sample it, and apply $\ell_2$ normalization:

\begin{align}
    \mathbf{f}_i^{\text{img}} = \operatorname{norm}_2\!\left(\text{bilinear}(\text{Conv}_{3\times3}(\text{DINOv2}(\mathbf{I})), \mathbf{u}_i)\right) \in \mathbb{R}^{384}.
\end{align}

For geometric features, we compute local point descriptors with a lightweight, multi-radius PointNet. Given the observed point cloud $\mathcal{P}=\{\mathbf{p}_i\in\mathbb{R}^3\}_{i=1}^N$, we aggregate neighborhoods at radii $r_1{<}r_2{<}r_3$ using radius grouping. We process each neighborhood with a small residual MLP followed by max pooling, then concatenate and fuse the resulting features:

\begin{align}
    \mathbf{f}_i^{\text{geo}} = \Phi(\oplus_{k=1}^3\max(\text{MLP}(\mathcal{N}_{r_k}(\mathbf{p}_i))))\in\mathbb{R}^{d_{\text{geo}}},
\end{align}

where $\oplus$ is concatenation and $\mathcal{N}_{r_k}(\mathbf{p}_i)$ represents the neighborhood points within radius $r_k$ of $\mathbf{p}_i$.

We project both modalities to a common width $d$ and concatenate:
\begin{align}
    \mathbf{f}_i = [\mathbf{W}_v\mathbf{f}_i^{\text{img}}\oplus\mathbf{W}_g\mathbf{f}_i^{\text{geo}}]\in\mathbb{R}^{2d},
\end{align}

\subsection{Network Architecture}
The network combines visual and geometric descriptors, encodes context-augmented point pairs, and predicts a joint NOCS--size flow. Each residual MLP block (ResLayer) contains two linear layers with ReLU between them and an additive skip connection; the skip uses a linear projection when the input and output widths differ.

\paragraph{Visual Branch}
We use DINOv2-S/14~\citep{oquab2023dinov2}, enhanced by 3DCorrEnhance~\citep{you2024multiview}, with rank-4 low-rank adapters on the query and value projections of its last four transformer blocks. The final normalized patch tokens form a 384-channel feature map. A $3\times3$ convolution with 384 input and output channels and padding 1 refines this map. Bilinear interpolation and $\ell_2$ normalization give per-point descriptors, which a linear layer projects from 384 to $d=512$ dimensions.

\paragraph{Geometric Branch}
We employ a multiscale PointNet~\citep{qi2017pointnet} at radii $\{0.02,0.04,0.08\}$, with neighborhood capacities of 32, 64, and 96 points. At each radius, three ResLayers map relative neighborhood coordinates through $3 \to 128 \to 128 \to 128$, followed by max pooling. The three pooled descriptors concatenate to 384 dimensions and pass through ResLayers of widths $384 \to 256 \to 256 \to 512$. A separate $512\to512$ linear projection produces the geometric descriptor. Concatenating it with the projected visual descriptor gives 1,024 dimensions per point.

\paragraph{Point Pair Feature Extraction}
Each target pair $\mathcal{T}=(i_1,i_2)$ is augmented with three context points, forming a five-point tuple $\mathcal{Q}_{\mathcal{T}}$. We concatenate all ten pairwise 3D differences in metric coordinates (30 dimensions). The five point descriptors concatenate to 5,120 dimensions and are projected linearly to 512. Their concatenation with the coordinate differences gives a 542-dimensional input to five ResLayers with widths $542\to512\to512\to512\to512\to512$. Four pre-normalized transformer encoder layers then exchange information across the sampled tuples, using eight attention heads and feed-forward width 2,048. Each target pair receives a 512-dimensional contextual feature.

\paragraph{Joint Flow Head}
The head models a nine-dimensional state $\mathbf{z}=[2\mathbf{x}_{i_1}\oplus2\mathbf{x}_{i_2}\oplus\log\mathbf{s}]$, comprising two NOCS points and three log box dimensions. A 512-dimensional sinusoidal time encoding passes through a $512\to512\to512$ MLP with ReLU between its linear layers. Concatenating the contextual feature, learned time embedding, and noisy state gives 1,033 inputs. Three ResLayers of widths $1033\to512\to512\to320$ predict 32 mixture components, each with a nine-dimensional flow mean and one mixture-weight logit. Log-softmax normalizes the logits.

A separate time-only MLP, with SiLU before each linear layer and widths $512\to512\to512\to1$, predicts a scalar log standard deviation shared across tuples, mixture components, and state coordinates. It receives a detached time embedding. After flow sampling, the first six state coordinates are divided by two to recover the target pair's NOCS coordinates, and the final three are exponentiated to recover metric box size. Thus both outputs come from the same joint head.

\subsection{Joint Flow Training}
\label{sec:joint_flow_training}
\paragraph{Targets and tuple sampling}
We sample all five tuple indices uniformly with replacement from the observed object points. The first two indices define the target pair, followed by the three context indices in descriptor concatenation. Target NOCS coordinates are clipped to $[-0.5,0.5]$ and multiplied by two; box dimensions are floored at $10^{-4}$ before taking their logarithms. Concatenating the two normalized NOCS points and three log box dimensions gives $\mathbf{z}_0\in\mathbb{R}^9$ for each tuple.

\paragraph{Forward bridge}
Let $\epsilon_t$ and $\rho$ denote the time floor and transition ratio. We draw $q\sim\mathcal{U}[0,1)$ per image and set
\begin{align}
    t&=\max(q,\epsilon_t), &
    \ell&=\max\!\left(0,\min\!\left((1-\rho)t,t-\epsilon_t\right)\right),
\end{align}
where $\ell=t_{\mathrm{low}}$. For the linear schedule $\alpha(t)=1-t$, $\sigma(t)=t$, two independent Gaussian vectors $\boldsymbol\epsilon_0,\boldsymbol\epsilon_1\sim\mathcal{N}(0,\mathbf{I}_9)$ construct the paired states:
\begin{align}
    \mathbf{z}_\ell&=(1-\ell)\mathbf{z}_0+\ell\boldsymbol\epsilon_0, \\
    \mathbf{z}_t&=a\mathbf{z}_\ell+\sqrt{t^2-a^2\ell^2}\,\boldsymbol\epsilon_1,
    \qquad a=\frac{1-t}{1-\ell}.
\end{align}
Each noise vector is shared across tuples within an image. The paired states have the stated noise marginals and are coupled through this forward bridge. The normalized time is multiplied by a fixed embedding scale before sinusoidal encoding.

\paragraph{Reverse-transition likelihood}
The network outputs flow-component means $\mathbf{u}_g\in\mathbb{R}^9$, weights $\pi_g$, and shared standard deviation $\sigma_u$. Following the Gaussian reverse bridge~\citep{gmflow}, define
\begin{align}
    b&=1-\left(\frac{\ell(1-t)}{t(1-\ell)}\right)^2, &
    c_1&=\left(\frac{\ell}{t}\right)^2\frac{1-t}{1-\ell}, \\
    c_2&=b(1-\ell), & c_3&=b\ell^2.
\end{align}
The induced transition mixture retains $\pi_g$ and has parameters
\begin{align}
    \mathbf{m}_g&=c_1\mathbf{z}_t+c_2(\mathbf{z}_t-t\mathbf{u}_g), &
    v&=(tc_2\sigma_u)^2+c_3.
\end{align}
We evaluate the mixture likelihood of the full nine-dimensional target $\mathbf{z}_\ell$ and average its negative logarithm over images and tuples. This jointly supervises NOCS and log-size through an analytical transition, allowing simulation-free training. For numerical stability, the reverse mapping floors denominators and log arguments at $10^{-6}$, and the likelihood caps inverse standard deviations at $10^4$.

\subsection{Implementation Settings}
We minimize $\mathcal{L}_{\mathrm{joint}}$ with $1{,}000$ point pairs per training image. We use AdamW with a base learning rate of $10^{-4}$, cosine decay with linear warmup, and gradient clipping at norm 1. Visual features come from DINOv2-S/14 enhanced by 3DCorrEnhance~\citep{you2024multiview}; the PointNet is trained from scratch with grouping radii $\{0.02,0.04,0.08\}$. We use a trainable geometric encoder to accommodate depth statistics across datasets and sensors; its feature contribution is evaluated in Section~\ref{sec:ablation}. Training uses the PACE, Omni6DPose, NOCS, and HouseCat6D training splits with balanced sampling. At inference, we sample $N{=}1{,}024$ object points and $T{=}10{,}000$ point pairs by default. Each pair uses an ODE sampler with 16 steps and 2 substeps per step. The pair-count trade-off is evaluated in Section~\ref{sec:ablation}.

\subsection{Training Details}
We train our model using the AdamW optimizer with a base learning rate of $10^{-4}$. We employ a cosine annealing learning rate schedule with linear warmup during the first epoch. The model is trained for 30 epochs on one NVIDIA A6000 GPU with a batch size of 16.
We use a balanced sampling strategy to ensure that each dataset (PACE, Omni6DPose, NOCS, HouseCat6D) contributes equally to the training batches.

\paragraph{Data Augmentation}
We apply data augmentation to improve generalization.
For RGB images, we apply random color jittering (brightness, contrast, saturation, hue) and add random Gaussian noise.
For point clouds, we apply random jittering to the points.
We also employ a Dynamic Zoom-In (DZI) strategy during training, where we randomly crop and resize the input images and point clouds to simulate different object scales and camera distances.

\section{Dataset Details}
In this section, we summarize the statistics of the training and test datasets used in our experiments.

\subsection{Training Datasets}
\paragraph{PACE}
We use the PACE-Sim training split of PACE~\citep{you2024pace}, which contains 100K photorealistic simulated frames with 2.4M annotations across 931 objects.

\paragraph{Omni6DPose}
We use the SOPE (Simulated Omni-Pose Estimation) split of Omni6DPose~\citep{zhang2024omni6dpose} for training. SOPE contains diverse objects placed in photorealistic simulated environments based on IKEA, Matterport3D, and ScanNet++ scenes. It provides 6D pose and metric-scale annotations for 475K images and 5M object instances, covering 4,162 objects from 149 categories.

\paragraph{NOCS REAL275}
We use the real training split of NOCS REAL275~\citep{wang2019normalized}, which contains approximately 4.3K real-world RGB-D frames from 7 scenes. It covers 6 object categories: bottle, bowl, camera, can, laptop, and mug.

\paragraph{HouseCat6D}
We use the training split of HouseCat6D~\citep{jung2024housecat6d}, which contains 34 large-scale scenes with approximately 20K real-world RGB-D frames captured in cluttered household environments. It covers 194 instances across 10 categories, including photometrically challenging classes such as glass and cutlery.

\subsection{Test Datasets}
\label{sec:mask_protocols}
Published baseline results retain their corresponding localization protocols, as summarized below.

\paragraph{PACE}
We use the official PACE test benchmark, which consists of 55K frames with 258K annotations across 300 videos, covering 238 objects from 43 categories. The published benchmark provides ground-truth detections to its category-level baselines, except NOCS, which jointly predicts detections and poses using Mask R-CNN~\citep{you2024pace}.

\paragraph{Omni6DPose}
For evaluation, we use the ROPE (Real Omni-Pose Estimation) split of Omni6DPose. ROPE consists of real-world video recordings of objects from the test set, providing a challenging benchmark for generalization to real sensor noise and lighting conditions. It contains 332K images with 1.5M annotations, covering 581 objects from 149 categories. Its published pose benchmark uses ground-truth instance masks for the two-stage methods; NOCS jointly predicts detections and poses~\citep{zhang2024omni6dpose}. The GenPose++ row in Table~\ref{tab:pose_comparison} is its reported ground-truth-mask result.

\paragraph{NOCS REAL275}
We evaluate on the real test split of NOCS REAL275, comprising 2.75K real-world frames from 6 unseen scenes, covering the same 6 categories. Published two-stage evaluations, including SecondPose and CleanPose, use predicted Mask R-CNN instance segmentations as inputs to pose estimation~\citep{chen2024secondpose,cleanpose2025}.

\paragraph{HouseCat6D}
We use the official 5-scene test split of HouseCat6D, which comprises approximately 3K real-world RGB-D frames covering the same 10 categories under unseen scene configurations and viewpoints. The original category-level pose benchmark evaluates its baselines using ground-truth detection masks~\citep{jung2024housecat6d}.

\paragraph{YCB-Video}
For cross-domain evaluation, we use the official test split of YCB-Video~\citep{xiang2017posecnn}, which consists of 12 RGB-D videos covering 21 YCB objects in cluttered tabletop scenes (the full benchmark contains 92 videos with 133,827 frames in total). Following standard practice, we evaluate on the keyframes of the 12 test videos. This dataset is excluded from the pose-training mixture. Dataset transfer and the stricter unseen-category subset evaluation are reported separately in the main paper.

\paragraph{HOPE}
For cross-domain evaluation, we use the HOPE-Image benchmark~\citep{tyree2022hope}, which provides RGB-D images annotated with 6-DoF poses of 28 toy grocery objects across 50 scenes from 10 household and office environments, with up to five lighting variations per scene. We evaluate on the 188 test images. This dataset is excluded from the pose-training mixture. Dataset transfer and the stricter unseen-category subset evaluation are reported separately in the main paper.

For YCB-Video and HOPE, Table~\ref{tab:in-the-wild} evaluates pose from provided instance masks/ROIs, reporting median rotation error and rotation/translation threshold accuracies.

\section{Baseline Results on Other Datasets}
\paragraph{Baseline Generalization}
Category-level methods, such as AG-Pose~\citep{lin2024instance}, VI-Net~\citep{lin2023vi}, and SecondPose~\citep{chen2024secondpose}, are constrained to the object categories on which they were trained (e.g., the six NOCS REAL275 categories). These methods are normally trained within a fixed taxonomy. We report a checkpoint-transfer diagnostic using their official NOCS REAL275 checkpoints on HouseCat6D. Table~\ref{tab:housecat6d_supp} reports their scores alongside our shared model. The training exposure differs: our model includes HouseCat6D training data, whereas these checkpoints do not. These numbers therefore do not isolate category generalization or architecture.

\begin{table}[!htbp]
\centering
\caption{\textbf{Baseline generalization results on HouseCat6D.} The \emph{Type} column matches the main paper: \textbf{Cat} = category-level specialist (per-dataset training, possibly with category labels or mean-shape priors), \textbf{Gen} = generalizable (no category labels, mean-shape priors, CAD models, or reference views at inference). The Cat baselines use NOCS REAL275 checkpoints; our model has seen HouseCat6D training data. This diagnostic has unequal training exposure.}
\label{tab:housecat6d_supp}
\resizebox{\linewidth}{!}{
\begin{tabular}{lccccccc}
\toprule
Method & Type
 & IoU$_{25}$ & IoU$_{50}$ &
 $5^\circ$2\,cm & $5^\circ$5\,cm & $10^\circ$2\,cm & $10^\circ$5\,cm \\
\midrule
VI-Net~\citep{lin2023vi}              & Cat  & 48.4 & 19.2 & 1.9  & 2.3  & 3.1  & 4.5 \\
AG-Pose~\citep{lin2024instance}       & Cat  & 79.6 & 38.9 & 0.8  & 1.5  & 2.2  & 4.5 \\
SecondPose~\citep{chen2024secondpose} & Cat  & 49.7 & 18.4 & 1.4  & 2.2  & 3.6  & 6.2 \\
\midrule
Ours                                 & Gen & \textbf{86.4} & \textbf{70.6} & \textbf{16.9} & \textbf{19.0} & \textbf{37.5} & \textbf{43.2} \\
\bottomrule
\end{tabular}
}
\end{table}

\section{N-hop Kabsch--Umeyama Pseudocode}
\paragraph{Discussion}
We combine point-pair-level context, multimodal pairwise NOCS prediction, and an adaptive N-hop estimator to accommodate partial observations and symmetry-induced ambiguity. Symmetric objects can still admit multiple equally valid poses. Pairwise scale calibration estimates the scalar normalization geometrically (Equation~\ref{eq:scale_cal}), while the learned metric box sizes determine the adaptive threshold and final box dimensions. The threshold can increase or decrease as the inlier set changes. The component ablation in Section~\ref{sec:ablation} evaluates their contributions to overall pose accuracy.

Algorithm~\ref{alg:nhop} fits a rigid transform to scale-calibrated correspondences. Here $\mathcal{J}$ indexes individual inlier correspondences, and $\operatorname{pairs}(\mathcal{J})$ selects each parent pair once if either endpoint is an inlier. Support and rigid refitting use individual correspondences; size aggregation uses the selected pairs. The loop assumes at least one supported hypothesis.

For observed points in meters, let $\mathbf{m}=\operatorname{med}_i\mathbf{p}_i$ and $\mathbf{a}=\operatorname{med}_i|\mathbf{p}_i-\mathbf{m}|$, using coordinate-wise absolute values and medians (lower medians for even sample counts). The extent proxy and initial threshold are
\begin{equation}
\label{eq:initial_threshold}
    d_{\mathrm{MAD}}=4\|\mathbf{a}\|_2,\qquad
    \tau_0=\max\!\left(0.01,\min\!\left(0.05,0.03d_{\mathrm{MAD}}\right)\right).
\end{equation}

\begin{algorithm}[!ht]
\caption{N-hop Kabsch--Umeyama with adaptive threshold}
\label{alg:nhop}
\DontPrintSemicolon
\KwIn{Flattened scale-calibrated correspondences $\mathcal{F}^{s}$ (two per pair), per-pair box sizes $\hat{\mathbf{s}}$, number of hops $H$, and initial threshold $\tau_0$}
\KwOut{$\mathbf{R}$, $\mathbf{t}$, $\mathbf{s}$}

$\tau := \tau_0$\;
$\mathcal{J}_{\text{best}} := \varnothing$\;
\For{$h = 1$ \KwTo $H$}{
    $(\hat{\mathbf{R}}, \hat{\mathbf{t}}, \mathcal{J}) :=\text{RANSAC-Umeyama}(\mathcal{F}^{s},\tau)$\;
    \If{$|\mathcal{J}| > |\mathcal{J}_{\text{best}}|$}{
        $\mathcal{J}_{\text{best}}:=\mathcal{J}$\;
    }
    \eIf{$|\mathcal{J}| > 0.1|\mathcal{F}^{s}|$}{
        $\tau := 0.05 \cdot \bigl\|\underset{\mathcal{T}_i\in\operatorname{pairs}(\mathcal{J})}{\operatorname{mean}}\hat{\mathbf{s}}_{\mathcal{T}_i}\bigr\|_\infty$\;
    }{
        $\tau := 1.5 \cdot \tau$\;
    }
}

$(\mathbf{R}, \mathbf{t}) := \text{Umeyama}(\mathcal{F}^{s}[\mathcal{J}_\text{best}])$\;
$\mathbf{s} := \underset{\mathcal{T}_i\in\operatorname{pairs}(\mathcal{J}_{\text{best}})}{\operatorname{mean}}\hat{\mathbf{s}}_{\mathcal{T}_i}$\;
\Return $\mathbf{R},\mathbf{t},\mathbf{s}$\;
\end{algorithm}

\clearpage
\section{Additional Qualitative Comparisons}
\begin{figure}[!ht]
    \centering
    \includegraphics[width=\linewidth]{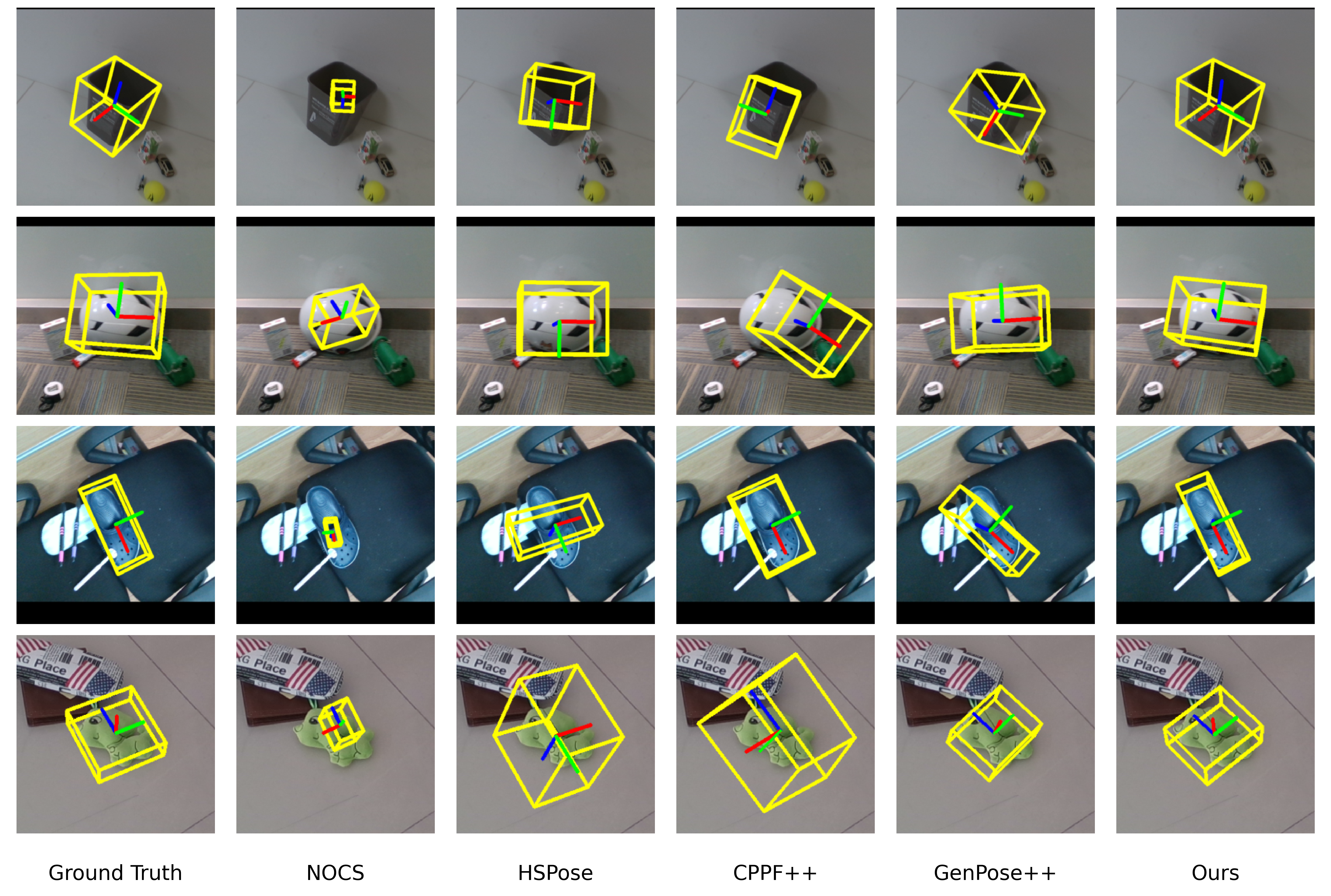}
    \caption{\textbf{Qualitative comparison on PACE.} Against NOCS, HS-Pose, CPPF++, and GenPose++, our method yields tighter 3D boxes and more consistent axes.}
    \label{fig:pace-vis}
\end{figure}

\begin{figure}[htbp]
    \centering
    \includegraphics[width=\linewidth]{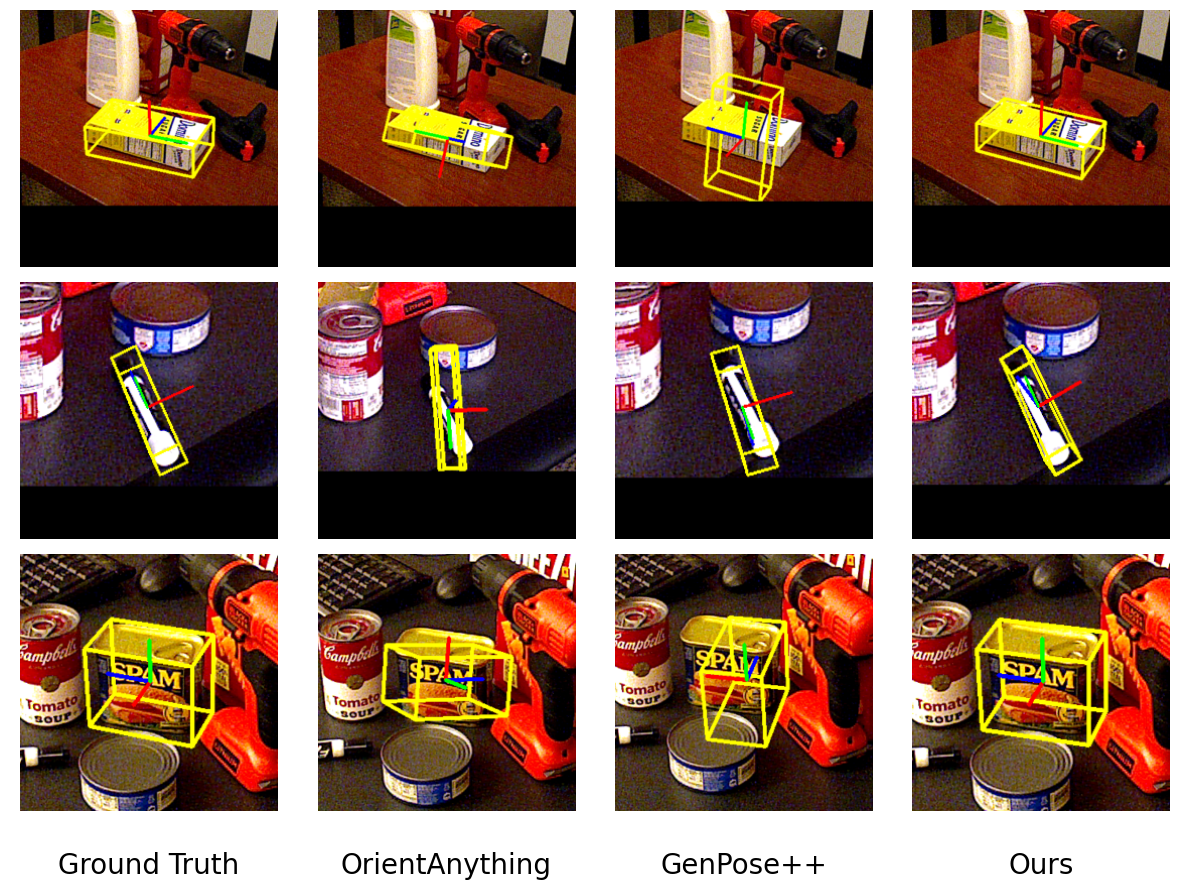}
    \caption{\textbf{Qualitative comparison on the YCB-Video dataset.} Compared with Orient Anything and GenPose++, our method estimates 9D poses for unseen objects in cluttered scenes with mutual occlusions.}
    \label{fig:supp_ycbv}
\end{figure}

\begin{figure}[ht]
    \centering
    \includegraphics[width=\linewidth]{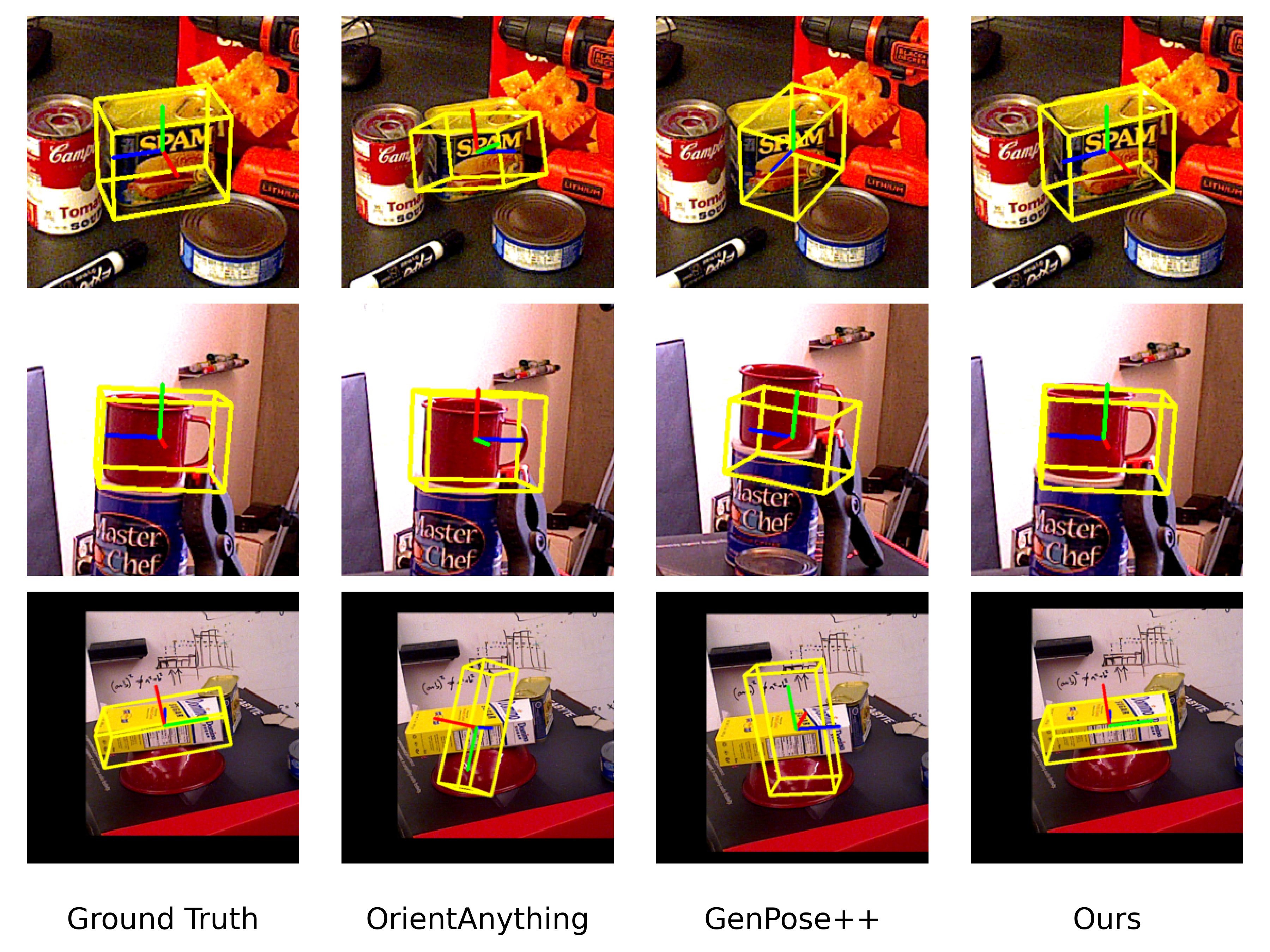}
    \caption{\textbf{YCB-Video qualitative comparison with RGB input.} In cluttered scenes with occlusion, our method produces tighter metric boxes and more stable axes than Orient Anything and GenPose++. We use the ground-truth translation and scale for the visualization of Orient Anything.}
    \label{fig:supp_ycbv_compare}
\end{figure}

\begin{figure}[ht]
    \centering
    \includegraphics[width=\linewidth]{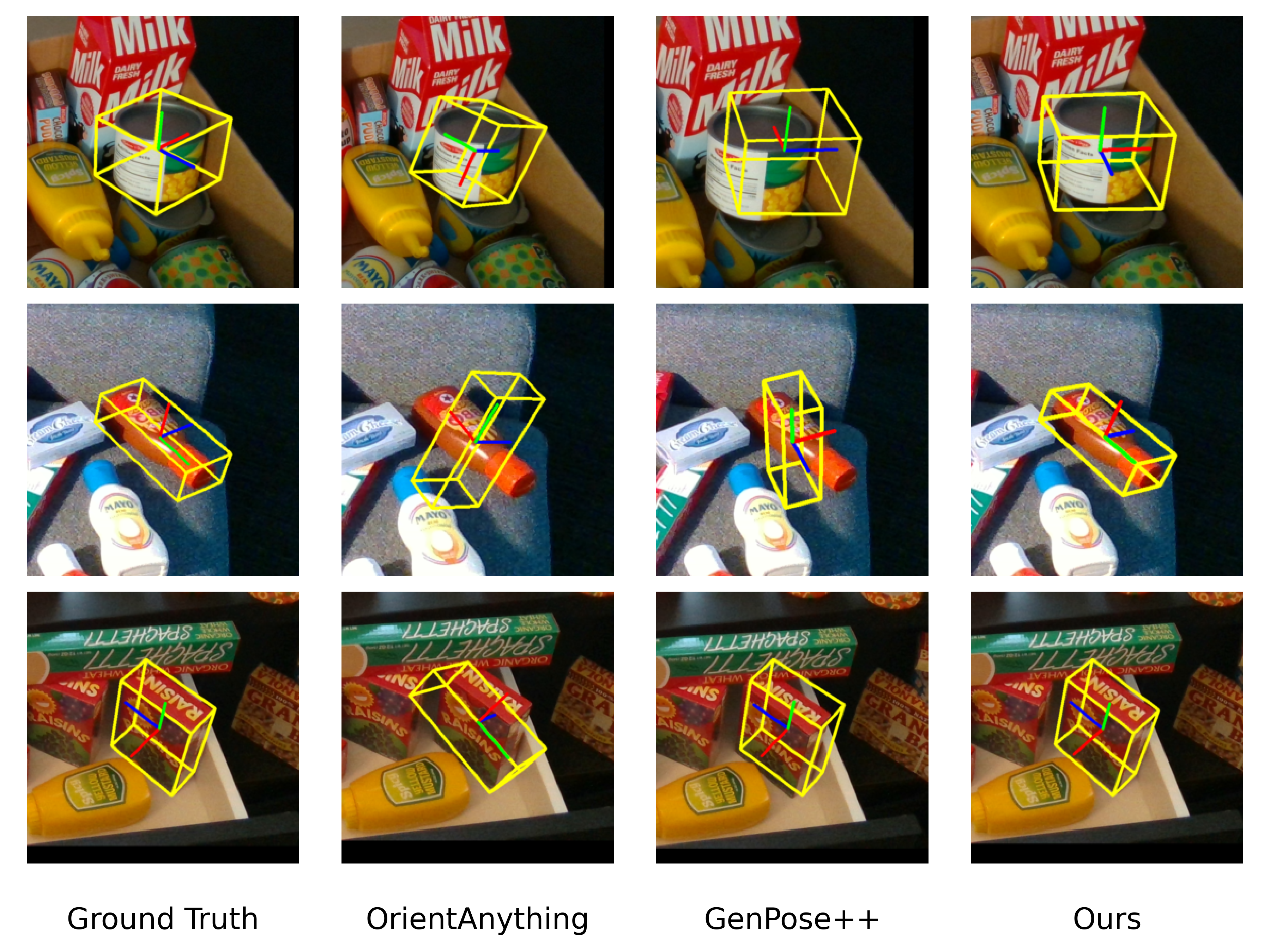}
    \caption{\textbf{HOPE qualitative comparison with RGB input.} In cluttered tabletop scenes with heavy occlusion and distractors, our method recovers tighter metric boxes and consistent axes than Orient Anything and GenPose++. We use the ground-truth translation and scale for the visualization of Orient Anything.}
    \label{fig:supp_hope_compare}
\end{figure}

\begin{figure}[ht]
    \centering
    \includegraphics[width=\linewidth]{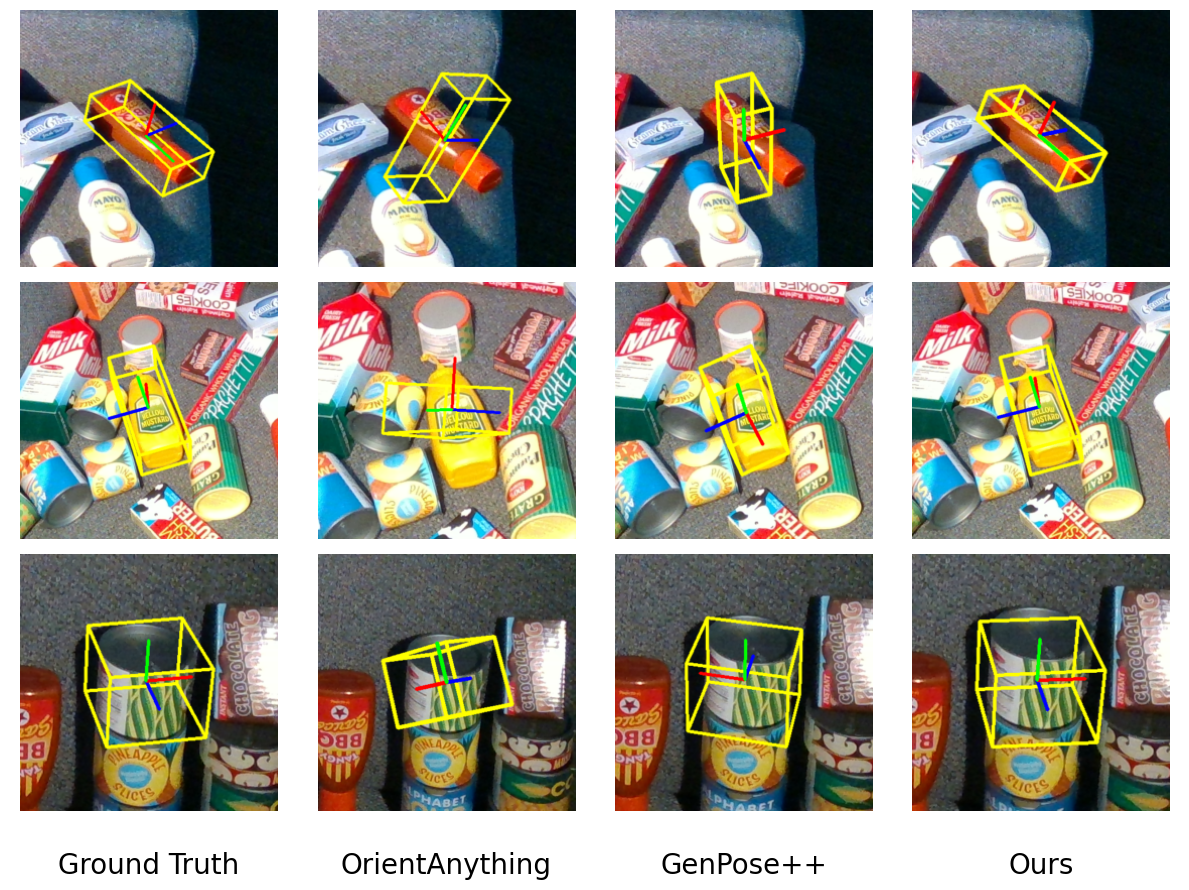}
    \caption{\textbf{Qualitative comparison on the HOPE dataset.} Compared with Orient Anything and GenPose++, our method recovers poses and metric box sizes despite heavy occlusions and diverse object types.}
    \label{fig:supp_hope}
\end{figure}

\begin{figure}[ht]
    \centering
    \includegraphics[width=\linewidth]{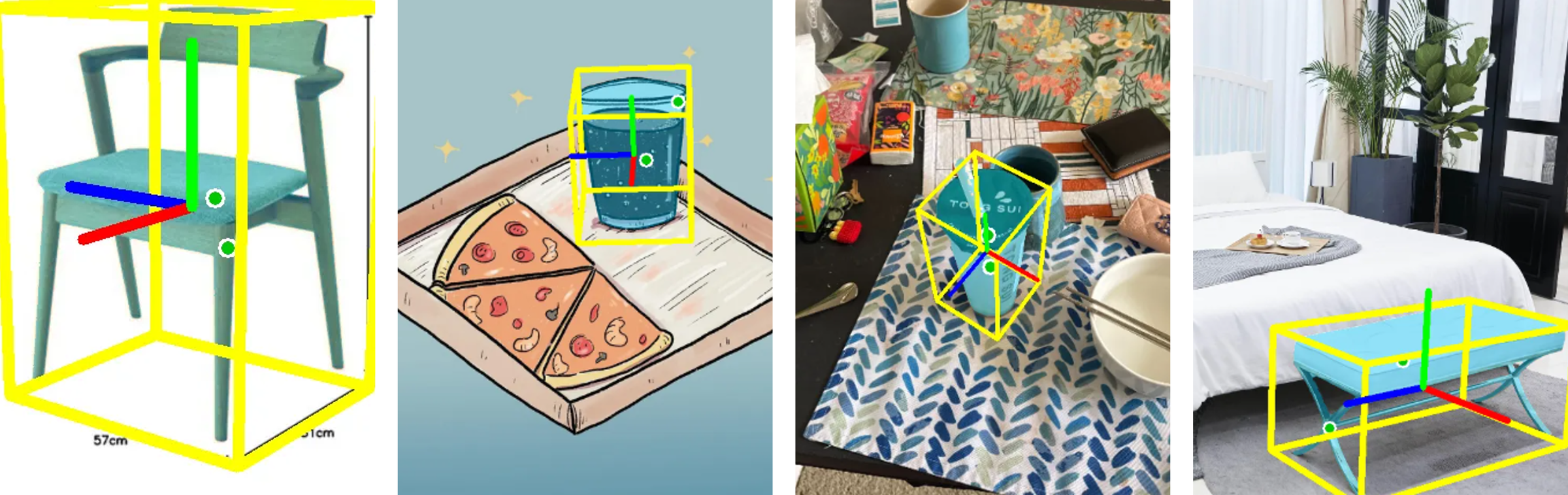}
    \caption{\textbf{Pose estimation in the wild.} Our method estimates 9D poses across images with artifacts and domain shifts.}
    \label{fig:supp_more}
\end{figure}

We present additional qualitative comparisons with Orient Anything and GenPose++ on the cross-domain YCB-Video dataset (Figures~\ref{fig:supp_ycbv} and~\ref{fig:supp_ycbv_compare}) and HOPE dataset (Figures~\ref{fig:supp_hope} and~\ref{fig:supp_hope_compare}). Both datasets contain diverse objects in cluttered scenes with severe occlusion. Our method recovers the 9D poses of these unseen objects without CAD models. Figure~\ref{fig:supp_more} shows results from additional image domains.

\clearpage
\section{Failure Cases and Limitations}
\begin{figure}[!htbp]
    \centering
    \includegraphics[width=\linewidth]{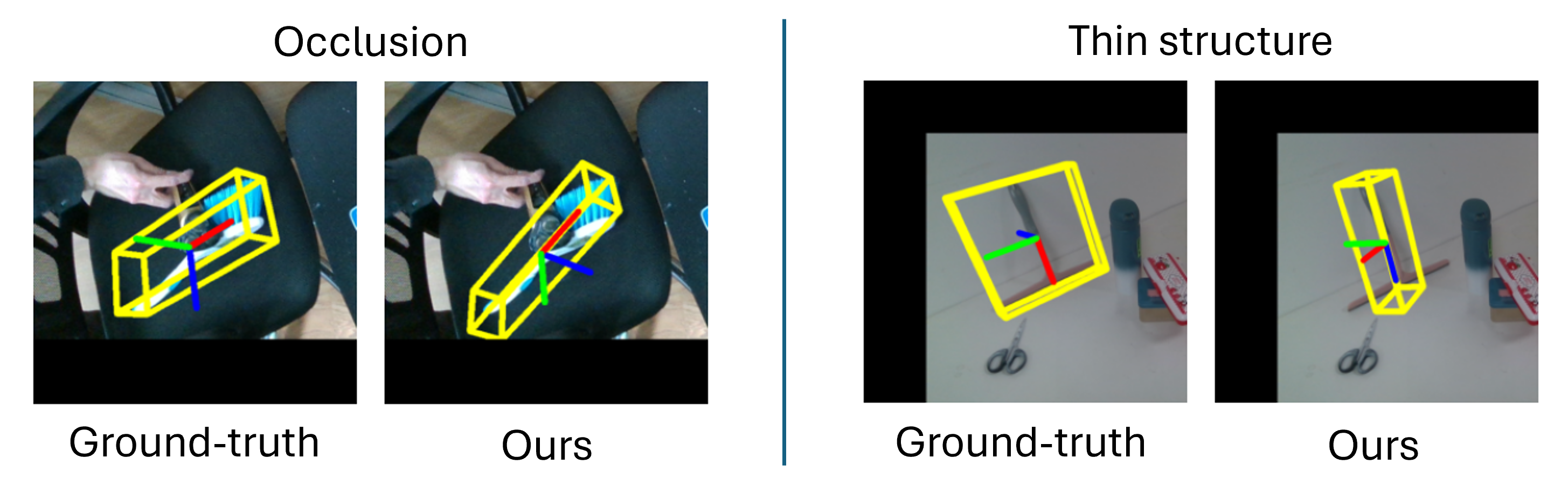}
    \caption{\textbf{Failure cases.} \textbf{Left:} Severe occlusion can produce unreliable NOCS predictions and incorrect pose estimates. \textbf{Right:} Very thin structures can create ambiguity in bounding box size and orientation.}
    \label{fig:limitation}
\end{figure}

Two visual failure modes are illustrated here; additional limitations are discussed in Section~\ref{sec:conclusion}.
First, as illustrated in the left panel of Figure~\ref{fig:limitation}, severe occlusion can hinder point-pair sampling and leave too few reliable correspondences for RANSAC to recover the correct pose.
Second, objects with extremely thin structures (right panel of Figure~\ref{fig:limitation}) can create ambiguity in the estimated bounding box size and orientation.